\documentclass{article} 
\usepackage[final]{colm2026_conference}

\usepackage{microtype}
\usepackage{url}
\usepackage{booktabs}
\usepackage{amsmath}
\usepackage{amssymb}
\usepackage{graphicx}
\usepackage{multirow}
\usepackage{enumitem}
\usepackage{bm}
\usepackage{hyperref}
\usepackage{cleveref}
\usepackage{xltabular}
\usepackage{subcaption}
\newcommand{\mathbm}[1]{\bm{#1}}
\usepackage[table]{xcolor}
\newcommand{\g}{\cellcolor{gray!12}}
\usepackage{caption}
\usepackage{tabularx}
\newcolumntype{Y}{>{\centering\arraybackslash}X}

\usepackage{listings}
\usepackage{lineno}

\definecolor{darkblue}{rgb}{0, 0, 0.5}
\hypersetup{colorlinks=true, citecolor=darkblue, linkcolor=darkblue, urlcolor=darkblue}

\title{When Large Language Models Know the Table: A Framework for Assessing Data Contamination in Tabular Datasets}

\author{Matteo Silvestri$^{1}$, Fabiano Veglianti$^{1}$ \\
\bf Flavio Giorgi$^{1,2}$, Fabrizio Silvestri$^{1}$, Gabriele Tolomei$^{1,2}$ \\
$^{1}$Sapienza University of Rome, Italy; \quad $^{2}$TellmewAI s.r.l. \\
Correspondence to \texttt{\{m.silvestri, gabriele.tolomei\}@uniroma1.it}
}

\begin{document}

\ifcolmsubmission
\linenumbers
\fi

\maketitle

\begin{abstract}
Large language models (LLMs) are increasingly exposed to \textit{data contamination}, i.e., performance gains driven by prior exposure of test datasets rather than generalization. However, in the context of tabular data, this problem is largely unexplored. Existing approaches primarily rely on memorization tests, which are too coarse to detect contamination.
In contrast, we propose a framework for assessing contamination in tabular datasets by generating controlled queries and performing comparative evaluation. Given a dataset, we craft multiple-choice aligned queries that preserve task structure while allowing systematic transformations of the underlying data. These transformations are designed to selectively disrupt dataset information while preserving partial knowledge, enabling us to isolate performance attributable to contamination. We complement this setup with non-neural baselines that provide reference performance, and we introduce a statistical testing procedure to formally detect significant deviations indicative of contamination.
Empirical results on eight widely used tabular datasets reveal clear evidence of contamination in four cases. These findings suggest that performance on downstream tasks involving such datasets may be substantially inflated, raising concerns about the reliability of current evaluation practices. 
\end{abstract}

\section{Introduction}

Large language models (LLMs) are now used across many tasks and domains \citep{survey-foundationmodels,survey-llmsapplications}, making reliable evaluation increasingly important. Benchmark results often guide deployment decisions and support claims of progress \citep{position-llmevaluation}. To be meaningful, these results must reflect a model’s ability to generalize to unseen data rather than familiarity with the evaluation set itself \citep{bias-llmevaluation}.

A major threat to this validity is \textit{data contamination} \citep{survey-datacontamination}, which we define here as overlap between training and test datasets. In this case, a model may perform well not because it learned the underlying task, but because it was previously exposed to the same dataset or highly similar instances during pretraining or post-training \citep{position-gptcontamination,position-nlpcontamination}. This can inflate benchmark scores and distort comparisons across models. Detecting contamination is therefore essential for trustworthy evaluation.\\
Contamination is related to, but distinct from, \textit{memorization}. Memorization asks whether a model stores and can reproduce training data, often in the context of privacy or data extraction \citep{memorization-carlini1}. Contamination, instead, is an evaluation problem: whether prior exposure to benchmark data renders performance an invalid measure of generalization \citep{bias-benchmarkcontamination}. A model need not reproduce test data verbatim for contamination to matter. What matters is whether it has non-trivial prior knowledge of the dataset that can be exploited to achieve unexpectedly strong performance.

This issue is especially relevant for \emph{tabular data}. LLMs are increasingly applied to structured inputs \citep{survey-llm4tabular}, in tasks such as table understanding \citep{llm-tableunderstanding}, data generation \citep{llm-datageneration}, and wrangling \citep{llm-datawrangling}, classification \citep{llm-classification1}, and reasoning over structured attributes \citep{llm-reasoning4tables}. At the same time, many tabular benchmarks are constructed from publicly accessible datasets \footnote{\url{https://www.kaggle.com/datasets} is a well-known source of tabular datasets} and are therefore plausible candidates for inclusion in model training corpora. Despite this risk, contamination in tabular settings remains poorly understood.\\
Existing approaches do not adequately address this problem. Prior contamination studies and detection frameworks have focused 
 on text benchmarks \citep{contamination-nlp,contamination-nlpquiz}, where the main signal is overlap between textual test examples and training data. By contrast, work on tabular data has mainly focused on verbatim memorization of individual rows \citep{contamination-tabularelephant}, rather than dataset-level contamination. This leaves a key methodological gap: there is still no principled framework to assess whether an LLM has prior knowledge of a public tabular dataset, nor a clear criterion for deciding when strong performance reflects contamination rather than ordinary task competence.

To address the need for reliable evaluation, we introduce a framework for assessing LLM contamination in tabular datasets using a dual-task multiple-choice benchmark. Unlike prior tabular approaches, which offer valuable memorization tests but no unified decision procedure, our framework grounds contamination assessment in a principled statistical testing procedure. Our method tests whether a model can recover masked attributes (completion) or verify an instance's presence (existence). By evaluating performance across dataset variants and comparing it against non-neural baselines, we can finely separate general or statistical knowledge from dataset exposure. While prior approaches proved too coarse, our framework successfully detected contamination in four of eight tested datasets, highlighting a concrete risk in LLM evaluation.
To summarize, our main contributions are:
\begin{enumerate}[label=\roman*), leftmargin=*]
    \item We introduce a framework for assessing contamination in tabular datasets used by LLMs. Given a dataset, we generate aligned multiple-choice queries across dataset variants to test a model's prior knowledge. Our code is publicly available at \url{https://github.com/hercolelab/tabular_contamination}.
    \item We propose non-neural baselines and a statistical testing procedure that provides a principled criterion for detecting contamination.
    \item We empirically show that prior memorization-based frameworks are too coarse to detect contamination across models and datasets, whereas our framework yields a finer-grained, more informative contamination profile.
\end{enumerate}

The remainder of the paper is structured as follows. \Cref{sec:related} reviews related work. \Cref{sec: framework} presents our framework, including the problem overview, dataset variants, evaluation probes, metrics, baselines, and result interpretation. \Cref{sec:experiments} describes the experimental setup and reports results for both prior memorization-based approaches and our method. \Cref{sec: discussion} discusses limitations and implications, and \Cref{sec: conclusion} concludes.

\section{Related Work}\label{sec:related}

LLMs can memorize training data \citep{memorization-carlini1,memorization-carlini3} and reveal stored content under suitable prompting conditions \citep{memorization-carlini2,llm-memorization}. This shows that training information can persist in model parameters, raising concerns that benchmark performance may be inflated by prior exposure to evaluation data \citep{position-nlpcontamination}.\\
This concern motivates the study of data contamination in benchmark evaluation \citep{bias-benchmarkcontamination}. In NLP, contamination has been studied through methods such as n-gram overlap, token distribution analysis \citep{contamination2-nlp,contamination-nlp}, and multiple-choice quizzes designed to test prior exposure \citep{contamination-nlpquiz}. These works show that benchmark leakage can substantially bias reported performance. Similar concerns arise in other domains: \citet{contamination-recsys} show that LLMs memorize MovieLens \citep{movielens}, a widely used recommendation benchmark, and that this prior knowledge can inflate evaluation results.

As LLMs are increasingly applied to tabular tasks like classification \citep{llm-classification1,llm-classification2,llm-classification3} and feature importance estimation \citep{llm-xai}, contamination in structured data has become a critical yet underexplored concern. Existing detection methods rely heavily on broad dataset knowledge or verbatim memorization \citep{contamination-tabularelephant,contamination-tabular2}; however, these signals are often too coarse to reliably indicate inflated benchmark performance (see \Cref{subsec:prior-framework-results}) or to help establish a standardized contamination registry, as claimed by \cite{position-nlpcontamination}. \citet{contamination-tabularelephant} rely on memorization tests that, as we show in \Cref{subsec:prior-framework-results}, fail to detect contamination in smaller and open-source models, limiting their reach across the current model landscape. \citet{contamination-tabular2} instead propose a decision rule for tabular contamination, but it reduces exposure to a binary flag rather than a statistically grounded test, leaving no room to express degrees of evidence. To overcome these limitations, we evaluate prior knowledge using multiple-choice quizzes derived from original tabular instances and controlled variants, comparing model behavior against non-neural baselines within a statistical testing procedure to yield robust, graded evidence of contamination.
Multiple-choice probing is itself an established paradigm in NLP contamination detection \citep{contamination-nlpquiz}, but transferring it to tabular data is not a direct port. Tables are governed by column-level marginal distributions, feature correlations, and heterogeneous attribute types, so a probe that ignores this structure cannot tell dataset exposure apart from generic distributional knowledge about the domain. Our contribution is therefore twofold: probes designed around this structure, and a guessing-baseline construction that turns their outcome into a statistically grounded decision criterion for contamination.

\section{Proposed Framework}\label{sec: framework}
\begin{figure}[t]
    \centering
    \includegraphics[width=1\linewidth]{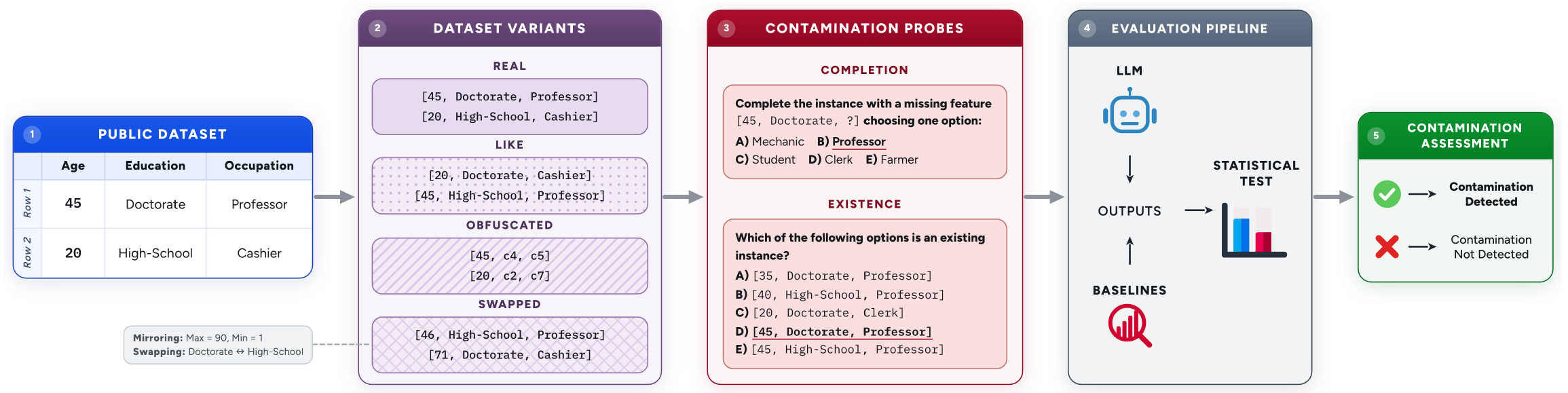}
    \caption{\textbf{Overview of the proposed framework for contamination assessment.} Given a public tabular dataset, we derive controlled dataset variants, construct contamination probes for completion and existence queries, evaluate LLMs against non-neural baselines, and use a statistical test to decide whether the observed behavior is indicative of contamination.}
    \label{fig:overview}
\end{figure}

In this section, we present our framework for assessing contamination in LLMs on tabular benchmarks (see \Cref{fig:overview}). Specifically, we focus on the problem overview, dataset variants, evaluation probes, metrics, baselines, and result interpretation.

\subsection{Problem Overview}\label{subsec:problem}
Because prior exposure to benchmarks threatens methodological integrity even without exact memorization, tabular datasets serve as an ideal testbed: their structured format limits verbatim recall while remaining semantically recognizable. To address this, our framework's goal is to investigate row-level contamination by evaluating an LLM's ability to recognize or reconstruct instance information under controlled transformations.

Let $D = \{r_1,\dots,r_N\}$ be a tabular dataset, whose rows share a fixed schema. We evaluate row-level contamination using five-way multiple-choice probes, specifically, \textit{completion} tasks (predicting masked cells) and \textit{existence} tasks (identifying an authentic row among distractors), to ensure cross-model comparability and establish explicit chance baselines. 
Ultimately, our framework assesses contamination by comparing this structured performance on the original dataset against transformed variants that isolate specific data characteristics.

\subsection{Dataset Variants}\label{subsec:dataset-variants}
For each dataset, we construct four variants: \texttt{real}, \texttt{like}, \texttt{swapped}, and \texttt{obfuscated}. 
These variants are generated from the same source table and are aligned so that the same underlying candidate rows are probed across conditions (more details in Appendix~\ref{app:dataset-variants}).

\noindent \textbf{Real.} The \texttt{real} variant is the unmodified cleaned dataset. It preserves the original feature names, original values, marginal distributions, and joint row-level dependencies. This is the reference condition in which contamination, if present, should be most directly expressed.

\noindent \textbf{Like.}
The \texttt{like} variant independently resamples each column from its empirical marginal distribution. It preserves univariate marginals but destroys row-wise dependence, so success here reflects column-level regularities rather than authentic row knowledge.

\noindent \textbf{Obfuscated.}
The \texttt{obfuscated} variant removes surface-level semantic cues. All feature names are renamed sequentially, e.g., to $f_1,f_2,\dots$, the target column is renamed to \texttt{target}, and categorical values are replaced with anonymous codes. This variant preserves row structure and value distributions while suppressing interpretable semantics, helping distinguish contamination from world knowledge or familiarity with the dataset name.

\noindent \textbf{Swapped.}
The \texttt{swapped} variant preserves schema and row structure but applies a within-column bijection so no value maps to itself. For categorical features, we map each element to another so that no value maps to itself; for numerical features with values in the interval $[N_{min}, N_{max}]$, we apply the map $\pi(x) := N_{max} - x - N_{min}$. This variant preserves row structure and column co-occurrence patterns.

\subsection{Task Generation}\label{subsec:tasks}
Tasks involve dataset-dependent queries that are aligned across dataset variants. We define a \textit{probe} as a structured multiple-choice query designed to test for dataset-specific knowledge. In the remainder of the paper, we use query and probe interchangeably to refer to these evaluation instances. This alignment ensures that differences in accuracy across variants are attributable to the transformation itself, rather than to changes in query difficulty. Details on probe construction and examples are provided in Appendix~\ref{app: probes_construction}.

\begin{figure}[t]
     \centering
     \begin{subfigure}{0.45\textwidth}
         \centering \includegraphics[width=\textwidth]{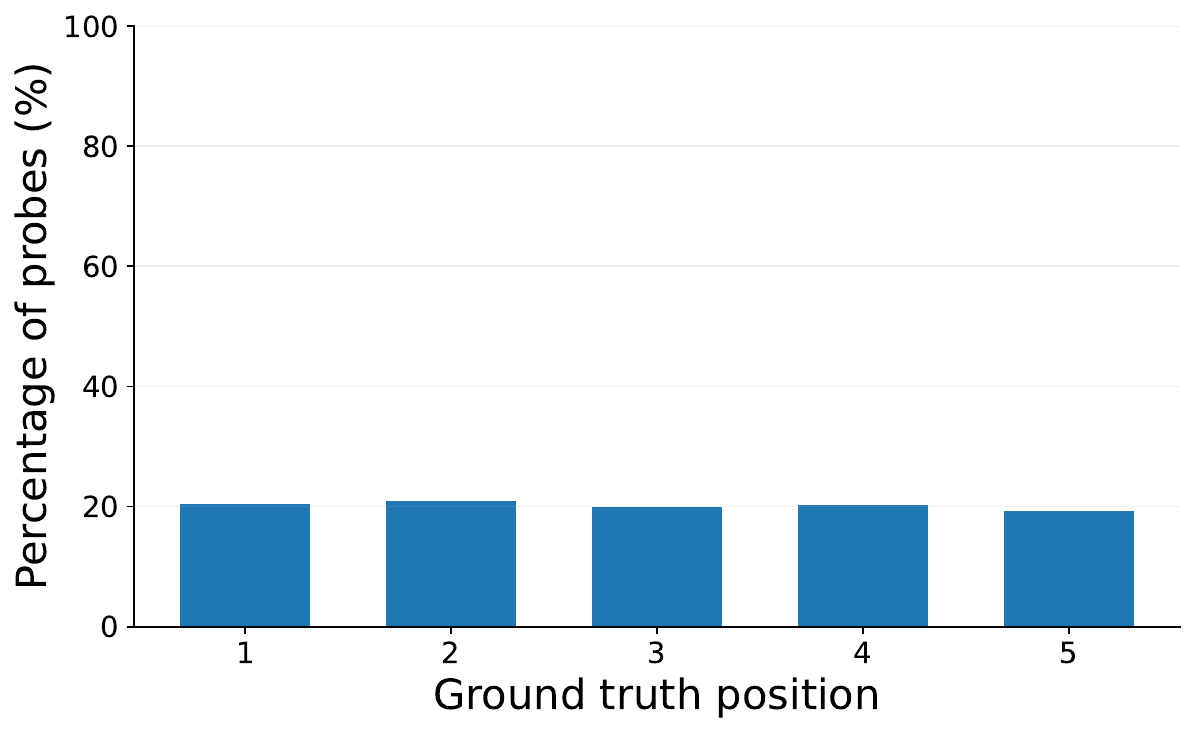}
         \caption{Indexes of correct answers in the probes.}
         \label{fig:position-probes}
     \end{subfigure}
     \hfill 
     \begin{subfigure}{0.45\textwidth}
         \centering \includegraphics[width=\textwidth]{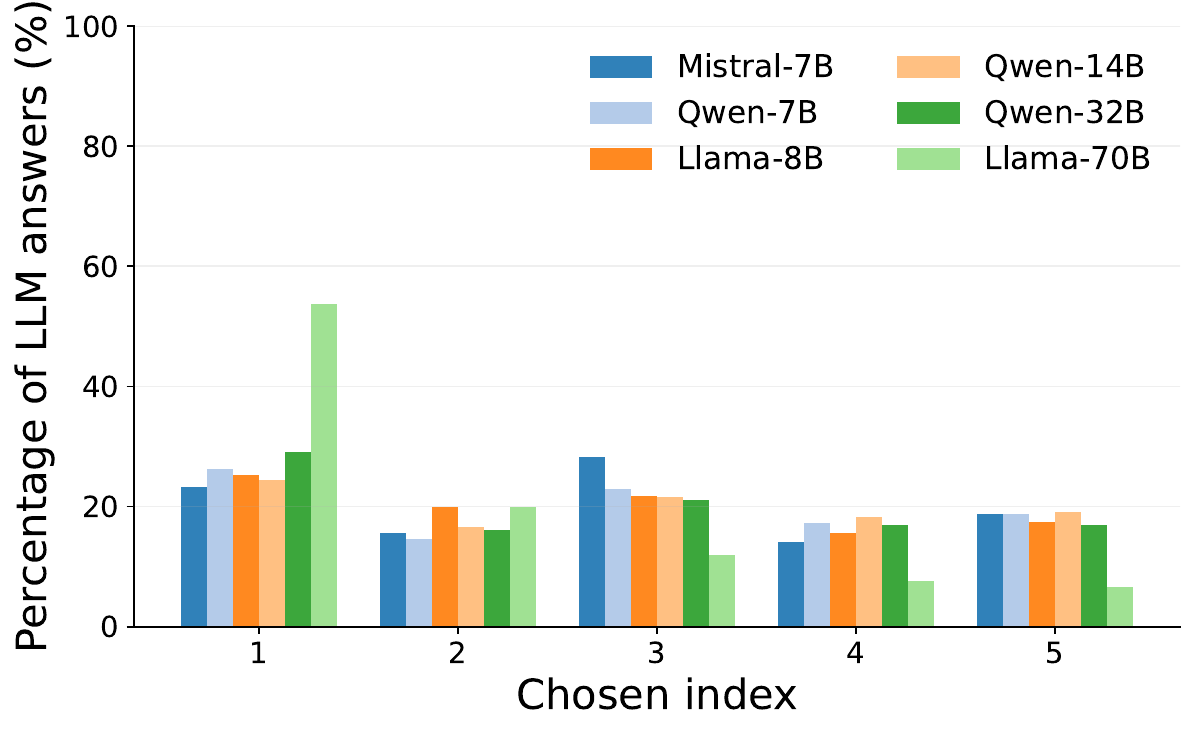}
         \caption{Chosen indexes by LLMs.}
         \label{fig:position-llm}
     \end{subfigure}
     
     \caption{\textbf{Analysis of positional bias}. We report the average position across datasets and its variants of the correct answers in the contamination probes (\Cref{fig:position-probes}), and the chosen indexes of LLMs during experiments (\Cref{fig:position-llm}).}
     \label{fig:position-answers}
\end{figure}

\noindent \textbf{Completion Task.} In a completion probe, a real row is partially observed, and the selected columns are replaced by question marks; then the model is asked to choose the correct values for the masked cells among five options. The correct option contains the true values, while the four distractors are sampled from the empirical supports of the masked columns and filtered to differ from the truth, from one another, and from original instances.
Specifically, let $M = \{c_{j_1},\dots,c_{j_m}\}$ denote the masked columns, then the prompt displays the row with $c_{j_\ell}=?$ for all $\ell$ and each option is formatted as a comma-separated assignment list, such as
\[
c_{j_1}=v_{j_1},\; c_{j_2}=v_{j_2},\; \dots,\; c_{j_m}=v_{j_m}.
\]
The position of the correct option in a prompt is randomly chosen. The random choice allows us to overcome potential position bias in the model (see \Cref{fig:position-answers}).
This task assesses whether the model can recover missing attribute values from the observed portion of a row. 

\noindent \textbf{Existence Task.}
In an existence probe, the model is shown five complete rows and asked to identify which one genuinely occurs in the dataset. Similarly to the completion task, the correct option is the true row, while the four distractors are generated by replacing the $M = \{c_{j_1},\dots,c_{j_m}\}$ target column values with alternative values sampled from the empirical support of the corresponding columns, ensuring that they differ from the true row and that distractors are unique.
The resulting options are fully serialized rows, and the five candidate rows are randomly shuffled. This task measures whether the model can distinguish authentic row-level combinations from plausible but fabricated near-matches.

\subsection{Metrics and Baselines}\label{subsec:metrics-baselines}
The framework is expressly targeted at contamination, hence it uses metrics and statistical tests against baselines to assert contamination.

\noindent \textbf{Evaluation Metrics.}
For a set of probes $\{1,\dots,n\}$, let $\hat{y}_i$ denote the model's chosen option index and $y_i$ the correct option index. The per-probe correctness indicator is $
z_i = \mathbb{I}[\hat{y}_i = y_i].
$
%

By denoting $\mathcal{C}$ and $\mathcal{E}$ the completion and existence probe subsets, respectively, and $|\mathcal{C}| = n_{\mathrm{comp}}$, $|\mathcal{E}| = n_{\mathrm{exist}}$, the framework reports task-specific accuracies:
\begin{equation*}
\mathcal{A}_{\mathrm{comp}} = \frac{1}{n_{\mathrm{comp}}}\sum_{i \in \mathcal{C}} z_i,
\qquad
\mathcal{A}_{\mathrm{exist}} = \frac{1}{n_{\mathrm{exist}}}\sum_{i \in \mathcal{E}} z_i,
\end{equation*}

Beyond absolute accuracy, the framework performs paired comparisons between the model and each baseline using McNemar's test \citep{mcnemartest}. Let $b$ denote the number of probes where the model is correct and the baseline is wrong, and $c$ the number where the model is wrong and the baseline is correct. The test is applied to these discordant pairs $(b,c)$.
We use a one-sided alternative
$
H_1: \{\text{model is better than baseline}\},
$
and compute separate $p$-values for completion, and existence probes.
This paired design is appropriate because both model and baseline are evaluated on the same set of probes, controlling for instance-level variability.

\noindent \textbf{Baselines.}
The framework includes three explicit baselines: a \textit{random} baseline, a  \textit{deterministic} baseline, and a  \textit{stochastic} baseline. The former selects one of the five options uniformly at random. This is the minimal reference level and captures pure chance performance under the multiple-choice protocol.
The deterministic and stochastic baselines rely on the probability distribution of options in a probe; specifically, the deterministic baseline selects the option with the highest probability, while the stochastic baseline samples one option from the distribution. Details on how the probability distribution of options in a probe is derived are provided in Appendix~\ref{app: baseline_prob_build}.

\subsection{Interpretation of the Framework}\label{subsec:interpretation}
Hereinafter, we consider a model contaminated if its statistical test against the random baseline is positive on \texttt{real}. This is a deliberate choice: we define contamination at the lowest detectable signal level, so that a significant test against the random baseline is read as evidence that the model has access to exploitable information about the original dataset beyond uninformed guessing, i.e., that the benchmark can no longer be treated as fully unseen data for that model. Beating the random baseline while failing to beat the deterministic or stochastic baseline carries weaker evidential weight than beating all three, but the two cases are not conflated: the random baseline is a necessary, not sufficient, condition for a stronger contamination claim, and success against the marginal-informed baselines signals a more pronounced form of contamination within this hierarchical interpretation of evidence.
Additionally, the framework supports a structured interpretation of model behavior. Strong performance on \texttt{real} but not on \texttt{like} indicates reliance on authentic row-level structure rather than marginals. Retained performance on \texttt{swapped} suggests robustness to categorical relabeling and reliance on relational structure. Retained performance on \texttt{obfuscated} rules out dependence on semantic cues (e.g., feature names or labels). Finally, a comparison with the deterministic and stochastic baselines shows how statistical information can be leveraged to solve the tasks.



\section{Experiments}\label{sec:experiments}

In this section, we present the experimental results. \Cref{subsec: experimental-setup} describes the experimental setup, including the evaluated models, datasets, and research goals. \Cref{subsec:results} then addresses these goals through the empirical analysis.

\subsection{Experimental Setup}\label{subsec: experimental-setup}


\noindent \textbf{Models.}
We evaluate seven large language models spanning different families and parameter scales. The open-weight models are Mistral 7B~\citep{model_mistral7b}, Qwen-\{7B, 14B,32B\}~\citep{model_qwen7b14b,model_qwen2.5_7b14b32b} and Llama-8B~\citep{model_llama3_8b}. In addition, we include GPT-OSS~\citep{model_gptoss} and Llama 70B~\citep{model_llama3_8b}. This model set allows us to study whether contamination signals are consistent across architectures, parameter scales, and model families.

\noindent \textbf{Datasets.}
We evaluate our framework on eight public tabular datasets: \texttt{adult}~\citep{dataset_adult}, \texttt{blood}~\citep{dataset_blood}, \texttt{credit}~\citep{dataset_credit}, \texttt{diabetes}~\citep{dataset_diabetes}, \texttt{gamma}~\citep{dataset_gamma}, \texttt{iris}~\citep{dataset_iris}, \texttt{mushroom}~\citep{dataset_mushrooms}, \texttt{titanic}\footnote{\url{https://www.kaggle.com/datasets/yasserh/titanic-dataset}}; these datasets span multiple domains and feature types. Additionally, we include a \texttt{synthetic} dataset as a controlled negative reference point.
Details on its construction are provided in Appendix~\ref{app:synthetic-dataset}.

\noindent \textbf{Framework Hyperparameters.}
We use 5 answer options per query and set the decoding temperature to 0 to ensure deterministic outputs. In the completion task, we mask $20\%$ of the columns, a trade-off between test difficulty and reliability of the estimated accuracy; \Cref{app:ablation} reports an ablation at $50\%$ coverage. Statistical significance for both completion and existence tests is set at $\alpha=0.01$. An ablation study on temperature and masking ratio is reported in the Appendix ~\ref{app:ablation}.

\noindent \textbf{Research Goals.}
Our experiments are organized around four primary research goals: (\textbf{RG1}) validating our framework's ability to reliably detect tabular dataset contamination in LLMs; (\textbf{RG2}) introducing a comprehensive contamination registry across various datasets and models; (\textbf{RG3}) evaluating whether the extent of contamination varies across different model scales; and (\textbf{RG4}) demonstrating that our approach reveals subtle contamination signals missed by existing, memorization-based approaches.

\subsection{Results}\label{subsec:results}
We first examine the limitations of existing memorization-based frameworks, then we present results from our framework, and summarize them in a contamination registry.

\subsubsection{Memorization-based Framework Results}
\label{subsec:prior-framework-results}

We compare our approach with the state-of-the-art framework of \cite{contamination-tabularelephant}, reporting all four of its memorization tests. \textit{Row completion} requires reconstructing a full row given $n_{\textrm{p}}$ preceding rows and $n_{\textrm{e}}$ few-shot examples, while \textit{feature completion} predicts a masked feature given the remaining features and $n_{\textrm{e}}$ examples. The \textit{header} test shows the model the first $n_{\textrm{p}}$ rows of the dataset and asks it to predict the next $m$ rows. Finally, the \textit{first token} (FT) test is identical to row completion, but accuracy is computed only on the first token of the completed row. We use the default hyperparameters.

\begin{table}[htpb]
\centering
\caption{Performance of the memorization-based tests of \cite{contamination-tabularelephant} across models and three famous datasets: \texttt{adult}, \texttt{iris}, and \texttt{titanic}.}
\renewcommand{\arraystretch}{1.2}
\resizebox{\linewidth}{!}{%
\begin{tabular}{llcccccc}
\toprule
\multirow{2}{*}{\textbf{Datasets}} & \multirow{2}{*}{\textbf{Task}} & \multicolumn{6}{c}{\textbf{Models}} \\ \cline{3-8} 
& & \textit{Mistral-7B} & \textit{Qwen-7B} & \textit{Llama-8B} & \textit{Qwen-14B} & \textit{Qwen-32B} & \textit{GPT-OSS} \\
\midrule
\multirow{4}{*}{\texttt{adult}} & Row & $0 / 250$ & $0 / 250$ & $0 / 250$ & $0 / 250$ & $0 / 250$ & $0 / 250$ \\
& Feature & $0 / 250$ & $0 / 250$ & $0 / 250$ & $0 / 250$ & $0 / 250$ & $0 / 250$ \\
& Header & $0 / 5$ & $0 / 5$ & $0 / 5$ & $0 / 6$ & $0 / 5$ & $0 / 0$ \\
& FT & $0 / 250$ & $0 / 250$ & $0 / 250$ & $0 / 250$ & $0 / 250$ & $0 / 250$ \\
\hline
\multirow{4}{*}{\texttt{iris}} & Row & $52 / 250$ & $109 / 250$ & $5 / 250$ & $1 / 250$ & $182 / 250$ & $48 / 250$ \\
& Feature & $2 / 250$ & $17 / 250$ & $6 / 250$ & $5 / 250$ & $9 / 250$ & $58 / 250$ \\
& Header & $0 / 16$ & $1 / 19$ & $1 / 17$ & $0 / 18$ & $3 / 17$ & $0 / 17$ \\
& FT & $0 / 250$ & $0 / 250$ & $0 / 250$ & $0 / 250$ & $0 / 250$ & $0 / 250$ \\
\hline
\multirow{4}{*}{\texttt{titanic}} & Row & $10 / 250$ & $0 / 250$ & $0 / 250$ & $10 / 250$ & $12 / 250$ & $0 / 250$ \\
& Feature & $10 / 250$ & $6 / 250$ & $2 / 250$ & $0 / 250$ & $3 / 250$ & $13 / 250$ \\
& Header & $0 / 19$ & $0 / 18$ & $0 / 19$ & $0 / 16$ & $0 / 19$ & $0 / 1$ \\
& FT & $0 / 250$ & $0 / 250$ & $0 / 250$ & $0 / 250$ & $0 / 250$ & $0 / 250$ \\
\hline
\end{tabular}%
}
\label{tab:comparison-sota}
\end{table}

\Cref{tab:comparison-sota} reports results on three widely used datasets. Memorization-based tests fail to detect contamination on canonical benchmarks such as \texttt{adult} and \texttt{titanic}, where all models achieve near-zero reconstruction accuracy. Given the prevalence of these datasets, this absence of detectable memorization is unexpected and is contradicted by the results in the next section.
More critically, these tests lack a criterion to determine when memorization implies contamination. This is evident on \texttt{iris}: some models (e.g., \textit{Qwen-32B}, 182/250 correct row completions) show clear memorization, while others (e.g., \textit{Llama-8B}, 5/250) do not. Their framework provides no statistical guidance to interpret such variability, leaving contamination ambiguous at both the dataset and model levels.
Overall, verbatim memorization is a coarse and unreliable proxy for tabular data contamination, underscoring the need for a more controlled, fine-grained framework.

\subsubsection{Proposed Framework Results}
\label{subsec:proposed-framework-results}

We analyze the results obtained with our framework. \Cref{fig:combined-tasks} illustrates representative behaviors, while complete results are reported in Appendix~\ref{app: additional-results}.


\noindent \textbf{Detecting contamination on \texttt{titanic}.}
The completion results on \texttt{titanic} (\Cref{fig:completion-titanic}) provide a clear example of the effectiveness of our framework. On the \texttt{real} variant, all models achieve accuracy significantly above the random baseline according to our statistical test. Under the decision rule introduced in Section~\ref{sec: framework}, this constitutes sufficient evidence of contamination for this dataset-model setting. Importantly, this addresses \textbf{RG1}, demonstrating that our framework can detect contamination in tabular datasets within LLMs.\\
The comparison across variants further clarifies the nature of the detected signal. Performance remains relatively high on  \texttt{like}, indicating that part of the advantage can be explained by knowledge of dataset-specific marginals. However, accuracy drops substantially on \texttt{swapped} and \texttt{obfuscated}, suggesting that the signal expressed on the \texttt{real} dataset is not merely due to generic tabular reasoning. Rather, the pattern is consistent with prior knowledge tied to the original dataset distribution and, at least in part, to its original feature-value semantics. Lastly, the comparison with the deterministic and stochastic baselines shows that the contamination effect is not solely due to statistical knowledge of the dataset.


\begin{figure}[htpb]
    \centering
    
    \begin{subfigure}[b]{1\textwidth}
        \centering
        \includegraphics[width=\linewidth]{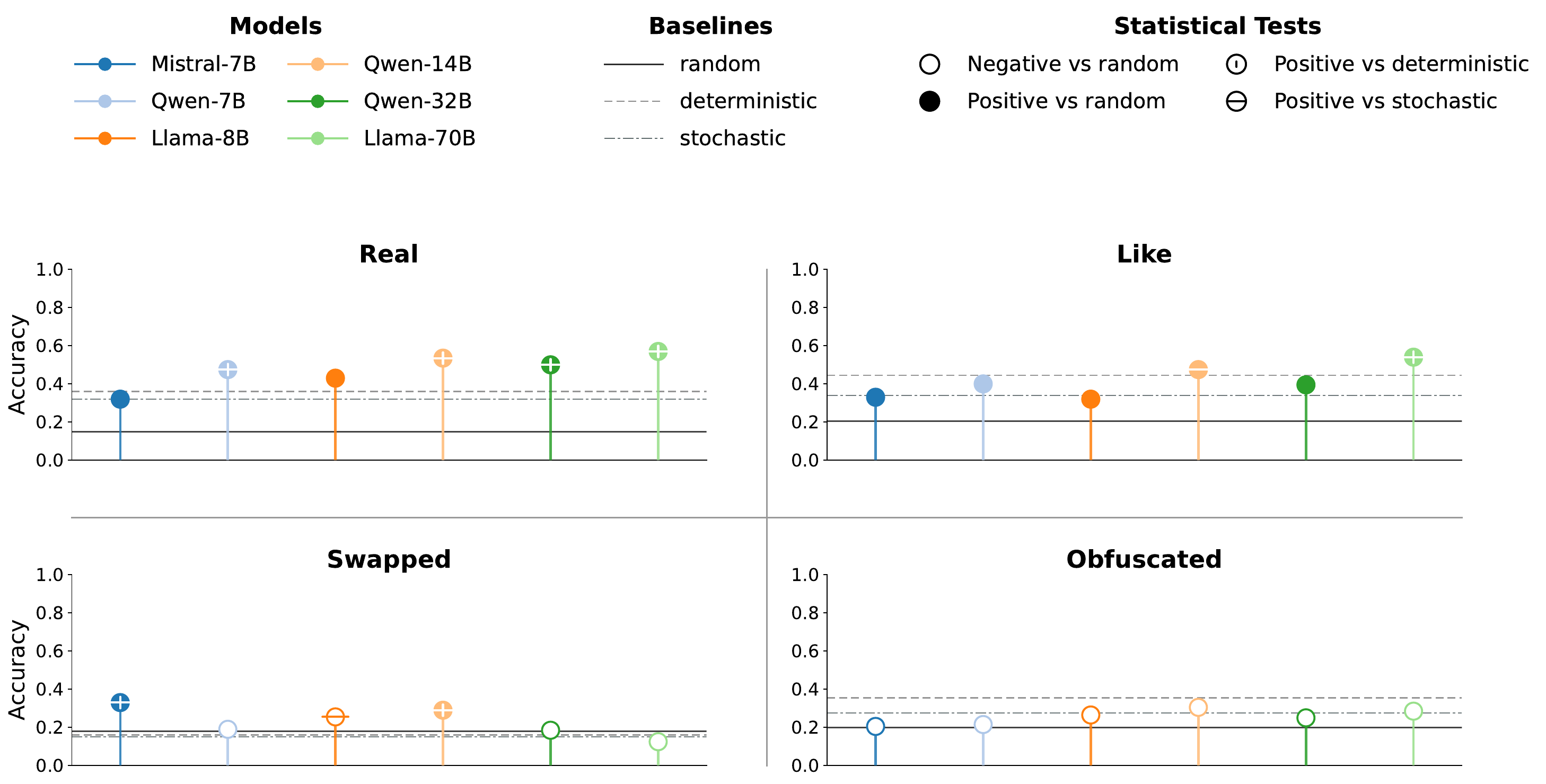}
        \caption{Completion of \texttt{titanic}}
        \label{fig:completion-titanic}
    \end{subfigure}
    
    \vspace{1.5em} 
    
    \begin{subfigure}[b]{1\textwidth}
        \centering
        \includegraphics[width=\linewidth]{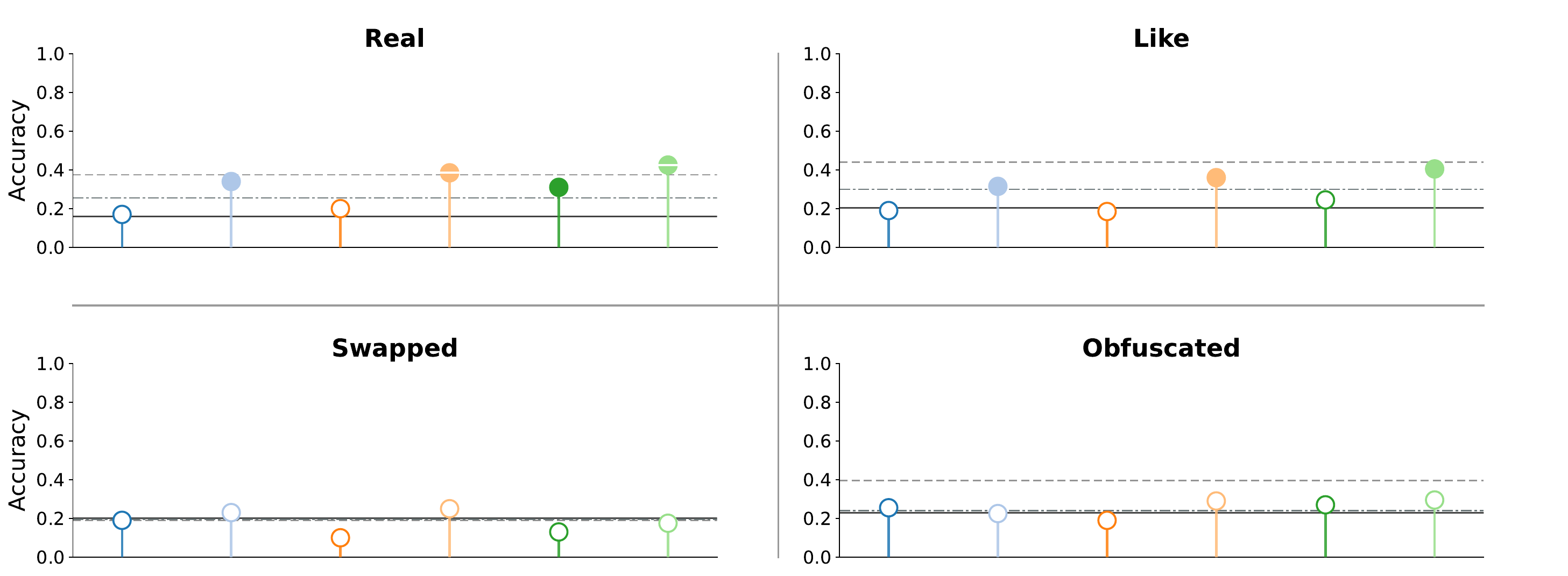}
        \caption{Existence of \texttt{titanic}}
        \label{fig:existence-titanic}
    \end{subfigure}
    
    \vspace{1.5em} 
    
    \begin{subfigure}[b]{1\textwidth}
        \centering
        \includegraphics[width=\linewidth]{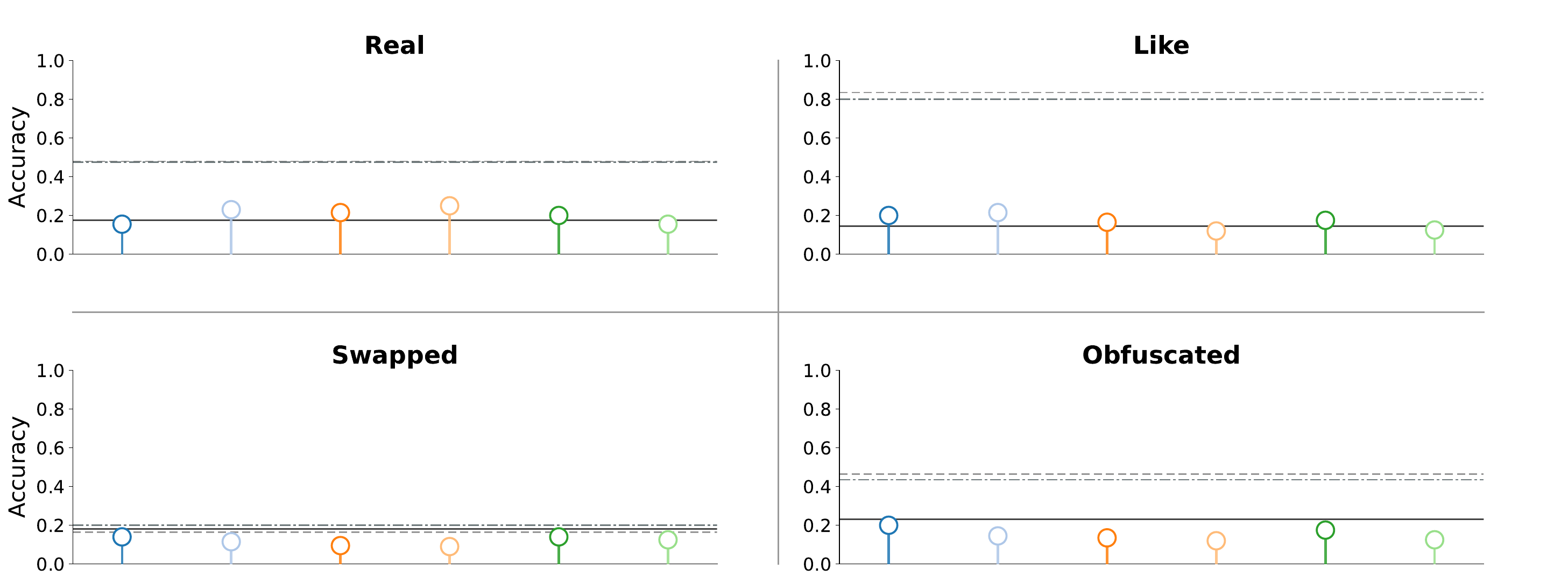}
        \caption{Completion of \texttt{mushroom}}
        \label{fig:completion-mushroom}
    \end{subfigure}
    
    \caption{\textbf{Task performance across dataset variants and models.} Baselines are represented as horizontal lines, and statistical test results using filled dots for detected contamination (positive against random), and horizontal/vertical bars on dots for superiority against statistical baselines.}
    \label{fig:combined-tasks}
\end{figure}

\noindent \textbf{Completion versus existence probes.}
The existence results on \texttt{titanic} (\Cref{fig:existence-titanic}) reveal a complementary pattern. In this case, only the larger models are significantly above the random baseline on the \textit{real} variant. This suggests that the existence task is intrinsically more difficult than the completion task. The same tendency is observed on other datasets (see \Cref{fig:real-transition}), indicating that completion probes are generally the more sensitive instrument for contamination detection.
At the same time, the existence task remains informative. Even though it is less sensitive, it still identifies contamination signals on \texttt{titanic} for the larger models, whereas the memorization-based framework reports near-zero row reconstruction accuracy on this dataset (see Section~\ref{subsec:prior-framework-results}). This is even more evident for the \texttt{adult} dataset (\Cref{fig:completion-adult,fig:existence-adult}). This addresses \textbf{RG4}: our framework captures contamination phenomena that are not revealed by existing methodologies.

\noindent \textbf{A negative case: \texttt{mushroom}.}
\Cref{fig:completion-mushroom} provides an equally important negative example. For \texttt{mushroom}, completion accuracy on the \texttt{real} variant remains close to chance and does not show a stable advantage over the transformed variants. In other words, our framework does not detect a contamination signal for this dataset. The result is important methodologically: it shows that the framework is not trivially flagging every public benchmark as contaminated, but can also return negative findings when the statistical evidence is insufficient. This is further validated by the \texttt{synthetic} dataset (see \Cref{fig:combined-synthetic}).

\noindent \textbf{Research Goals.}
Taken together, these results address all the research goals of this section. First, the \texttt{titanic} case demonstrates that our framework can reliably detect contamination, thereby addressing \textbf{RG1}. Second, the full set of results in Appendix~\ref{app: additional-results} naturally supports the construction of a dataset-model contamination registry, depicted in \Cref{tab:model-datasets}, which addresses \textbf{RG2}. Third, a clear scale-related trend emerges from \Cref{fig:completion-titanic,fig:existence-titanic} and from the additional results in the appendix: contamination signals generally become stronger as model size increases. 
This directly addresses \textbf{RG3} and suggests that larger models are more likely to express the consequences of prior dataset exposure. Finally, the contrast between our results and those of the memorization-based framework shows that contamination signals can be present even when verbatim reconstruction fails, addressing \textbf{RG4}.

\begin{table}[h]
\centering
\caption{\textbf{Contamination registry across datasets and models.} A checkmark ($\checkmark$) indicates that our framework detects contamination for the corresponding dataset–model pair, while a cross ($\times$) indicates no detected contamination under our evaluation protocol. The results highlight contamination consistently detected on \texttt{titanic}, \texttt{adult}, \texttt{credit}, and \texttt{iris}.}
\renewcommand{\arraystretch}{1.15}
\resizebox{0.9\linewidth}{!}{%
\begin{tabular}{lcccccc}
\hline
\multirow{2}{*}[-0.2ex]{\textbf{Datasets}}
& \multicolumn{6}{c}{\textbf{Models}} \\[0.4ex]
\cline{2-7}
& \textit{Mistral-7B}
& \textit{Qwen-7B}
& \textit{Llama-8B}
& \textit{Qwen-14B}
& \textit{Qwen-32B}
& \textit{Llama-70B} \\
\hline
\texttt{adult}     & $\checkmark$ & $\checkmark$ & $\checkmark$ & $\checkmark$ & $\checkmark$ & $\checkmark$ \\
\texttt{blood}     & $\times$ & $\times$ & $\times$ & $\times$ & $\times$ & $\times$  \\
\texttt{credit}    & $\times$ & $\checkmark$ & $\checkmark$ & $\checkmark$ & $\checkmark$ & $\checkmark$  \\
\texttt{diabetes}  & $\times$ & $\times$ & $\times$ & $\times$ & $\times$ & $\times$  \\
\texttt{gamma}     & $\times$ & $\times$ & $\times$ & $\times$ & $\times$ & $\times$  \\
\texttt{iris}      & $\times$ & $\checkmark$ & $\times$ & $\checkmark$ & $\checkmark$ & $\checkmark$  \\
\texttt{mushroom}  & $\times$ & $\times$ & $\times$ & $\times$ & $\times$ & $\times$  \\
\texttt{synthetic} & $\times$ & $\times$ & $\times$ & $\times$ & $\times$ & $\times$  \\
\texttt{titanic}   & $\checkmark$ & $\checkmark$ & $\checkmark$ & $\checkmark$ & $\checkmark$ & $\checkmark$  \\
\hline
\end{tabular}%
}

\label{tab:model-datasets}
\end{table}

\section{Discussion and Limitations}\label{sec: discussion}

Our results show that tabular data contamination in LLMs is a real methodological concern, challenging the common assumption that public tabular benchmarks are unseen at evaluation time. If a model has prior knowledge of a benchmark, evaluation reflects both generalization and exposure, inflating scores and weakening claims of progress. Crucially, this concern does not depend on observing a measurable gain: contamination is problematic even when it yields no detectable increase in downstream accuracy, because it already violates the assumption that benchmark examples are unseen at evaluation time. The primary harm is the loss of the guarantee that makes benchmark numbers interpretable, not the size of the inflation they exhibit.
We therefore argue that contamination assessment should be standard in studies evaluating LLMs on public tabular datasets. Contaminated datasets need not be discarded, but their use should be reported transparently and their results interpreted as potentially biased upward, and claims about model generalization should be correspondingly softened. Like dataset documentation or significance testing, contamination assessment does not replace evaluation; it makes it interpretable. A second implication concerns model scale: contamination tends to be more visible in larger models, suggesting that they are better able to exploit prior exposure. As a result, contamination can bias not only absolute performance estimates but also comparisons across model sizes, making scale-related gains appear larger than they would on truly unseen data. This is particularly important because benchmark leaderboards are often used to support claims about the superiority of larger models. Our results indicate that part of this superiority may, in contaminated settings, reflect a greater ability to exploit contamination rather than a cleaner improvement in generalization. 

We acknowledge that our work has some limitations. First, it \emph{can be used to show the presence of data contamination, but never to show its absence}. A negative result in our framework must be interpreted as an absence of detected evidence, not as evidence that the model has never been exposed to the dataset. This is especially relevant for datasets such as \texttt{mushroom}, where we detect no contamination despite the dataset being old and publicly available. A second limitation is sensitivity: although our framework is more sensitive than prior approaches, it may still miss contamination. This is plausible for old, widely circulated benchmarks and suggests that contamination detection remains, in part, a problem of test sensitivity. We conjecture that contamination detectability is inversely related to probe difficulty: verbatim memorization is the hardest and least sensitive signal, row existence is intermediate, and row completion is the easiest and most sensitive in practice. This interpretation is consistent with our coverage ablation: higher coverage makes row completion probes harder and contamination harder to detect. A more principled way to calibrate the framework's sensitivity and specificity would be to establish ground-truth contamination through controlled injection, e.g., via continued pretraining on post-release or synthetic tabular datasets.
Future work should therefore investigate even more sensitive protocols, calibrated against such ground-truth contamination, while preserving enough control to avoid false positives driven by superficial dataset regularities. A third limitation concerns the granularity of the knowledge that contamination induces. Even when a model is contaminated, it is not obvious \emph{what} exactly the model knows about the dataset. Our crafted variants provide a first step toward discovering the kind of knowledge.
However, this analysis is still limited. It does not yet fully characterize which parts of the dataset the model is confident about, which attributes are more strongly affected by contamination, or how contamination interacts with task formulation. A more fine-grained understanding of contaminated knowledge remains an important direction for future work.
A final limitation is that we do not quantify how the detected contamination translates into inflated downstream performance. Our probes establish that dataset knowledge is present, not how much of a benchmark score it accounts for. A systematic study of downstream inflation requires a dedicated controlled setup, and must account for the fact that leakage need not surface as higher accuracy if the retained information is not elicited by the prompting and evaluation protocol in use. We leave this to future work.

\section{Conclusion}\label{sec: conclusion}

In this work, we propose a framework for assessing contamination in tabular datasets by crafting multiple-choice queries and performing comparative evaluation using dataset variants and non-neural baselines.
The main message of this paper is twofold. First, contamination in tabular benchmarks is real and methodologically consequential, so evaluations that ignore it risk overstate both absolute performance and relative differences between models. Second, the contamination assessment itself remains an open problem: our framework provides a useful, more sensitive tool, but not a complete solution. We believe that this work will motivate the community to treat contamination analysis as a necessary part of rigorous LLM evaluation, and to develop stronger methods for identifying not only whether a model has seen a dataset before, but also which properties of that dataset continue to influence its behavior.

%
% ...

\section*{Ethical Statement}

This work does not involve personal data or sensitive information. All experiments are conducted on publicly available benchmark datasets commonly used in prior work. We do not identify any direct ethical risks or broader societal harms arising from this study. On the contrary, by improving the rigor of contamination detection, our work contributes to a more reliable and transparent evaluation of large language models.

\section*{Acknowledgments}

This work was partially supported by the following projects: GHOST – Protecting User Privacy from Community Detection in Social Networks (B83C24007070005) and SEEDS – Sustainable Ecosystems in Evolving Digital Societies, funded by Sapienza University of Rome through the ``Progetti di Ricerca Grandi" and ``Progetti Dipartimentali" programs, respectively.

\bibliography{references}
\bibliographystyle{colm2026_conference}

\appendix

\section{Details of Probes and Datasets Construction}

\subsection{Probes Construction}\label{app: probes_construction}
A key methodological aspect is that candidate rows and masking patterns are \emph{aligned across variants}. The same underlying row indices are selected for the \texttt{real}, \texttt{like}, \texttt{swapped}, and \texttt{obfuscated} conditions, and the same masked column subsets are reused after mapping names appropriately. This paired construction is crucial because it ensures that differences in accuracy across variants are attributable to the transformation itself rather than to changes in probe difficulty.
For each dataset variant, the framework generates $400$ probes in total, split equally between completion and existence tasks.

\subsubsection{Row Selection}
Probe generation is not uniform over rows. Instead, the framework preferentially samples \emph{rare} or distinctive rows, which are more informative for contamination analysis. Each row receives a rarity score computed from inverse empirical frequencies over randomly selected small subsets of columns. Specifically, in several trials, the algorithm selects a small subset of columns and computes an inverse-frequency product for each row; the final rarity score is the average of the log-transformed products across trials.

Intuitively, rows composed of uncommon value combinations receive higher scores. The generator ranks rows by this rarity measure, retains a pool of highly ranked candidates, shuffles that pool, and then constructs probes from it. This design biases the benchmark toward rows whose correct identification is less likely to be explained by coarse statistical regularities alone, and strengthens the reliability of the contamination-detected cases.

\subsubsection{Column Selection}
The framework allows evaluating contamination at different \emph{coverage} levels (e.g., $20\%$, $50\%$), corresponding to the proportion of non-target columns that are involved in the probe.
Let $p$ denote the number of non-target columns in the dataset. The number of masked or mutated columns, denoted $m$, is computed as
\begin{equation*}
m = \max\!\left(1,\min\!\left(p,\left\lfloor p \cdot \mathrm{coverage} + 0.5 \right\rfloor\right)\right).
\end{equation*}
Hence, coverage is operationalized as a rounded proportion of the available non-target attributes.
The target column is excluded from masking. This choice is methodologically important: it prevents the task from degenerating into a direct label-prediction problem and keeps the focus on contamination of feature configurations and row identity.

For each probe, the set of masked or mutated columns is chosen with a mixed strategy intended to cover both categorical and numerical information:
\begin{itemize}
    \item categorical columns are preferred according to empirical entropy, with a preference for non-binary variables when possible;
    \item numerical columns are preferred according to variance;
    \item when only one column is selected, binary categorical variables are avoided if non-binary or numerical alternatives exist.
\end{itemize}

This procedure avoids trivially easy probes and encourages questions that depend on informative attributes.

\subsection{Dataset Variants}\label{app:dataset-variants}

\noindent \textbf{Real.} The \texttt{real} variant is the unmodified cleaned dataset. It preserves the original feature names, original values, marginal distributions, and joint row-level dependencies. This is the reference condition in which contamination, if present, should be most directly expressed.

\noindent \textbf{Like.}
The \texttt{like} variant is obtained by independently resampling each column from its empirical marginal distribution. Formally, if the original dataset has columns $c_1,\dots,c_p$, then the transformed dataset $D^{\mathrm{like}}$ is generated by sampling each entry of column $c_j$ independently from the observed non-missing values of $c_j$ in $D$. 
This transformation preserves univariate marginals but destroys row-wise joint structure. Accordingly, success on this condition reflects knowledge compatible with column-level regularities, but not necessarily contamination of authentic rows.

\noindent \textbf{Obfuscated.}
The \texttt{obfuscated} variant removes surface-level semantic cues. All feature names are renamed sequentially, e.g., to $f_1,f_2,\dots$, with the target column, when known, renamed to \texttt{target}. Categorical values are also replaced by anonymous codes such as \texttt{c0}, \texttt{c1}, and so on. For binary targets, the target is converted to a binary encoding, typically $\{0,1\}$.
Thus, the obfuscated condition preserves the row-level structure and the distributional pattern of values, but suppresses human-interpretable semantics associated with feature names, categorical feature values, and category labels. This is a particularly stringent condition for distinguishing contamination from ordinary world knowledge or familiarity with the dataset name. Note that this variant has a stronger effect on datasets composed entirely of categorical features.

\noindent \textbf{Swapped.}
The \texttt{swapped} variant preserves the table schema but applies a strict within-column derangement. 
For each categorical feature, the set of unique observed values is permuted so that no category maps to itself. If a column has categories $\mathcal{V}_c$, the method constructs a bijection
\begin{equation*}
\pi_c : \mathcal{V}_c \to \mathcal{V}_c
\end{equation*}
such that $\pi_c(v) \neq v$ for every $v \in \mathcal{V}_c$.
For numerical features whose values fall within the interval $\mathcal{I}_n:=[N_{min}, N_{max}]$, the bijection used is
\begin{equation*}
        \pi_n:\mathcal{I}_n\rightarrow\mathcal{I}_n\\, \quad \pi_n(x) := N_{max} - x + N_{min}
\end{equation*}

This operation preserves the row structure and the pattern of co-occurrence across columns.

\subsection{Synthetic Dataset}\label{app:synthetic-dataset}
We construct a synthetic dataset as a negative control to validate that the framework does not detect contamination when contamination is absent. By design, this dataset has no prior exposure in LLM training data; thus, any detected signal would indicate a failure mode of the method. Importantly, the framework is not intended to certify the absence of contamination, but to provide evidence of its presence.

The dataset is generated deterministically (fixed random seed set to 42) using \textit{make\_classification}\footnote{\url{https://scikit-learn.org/stable/modules/generated/sklearn.datasets.make_classification.html}}, a predefined function of \textit{scikit-learn}, with 50{,}000 samples, 10 initial continuous features, and a binary target (class imbalance 75/25, 1\% label noise, moderate separability). Five categorical features are then derived from the continuous space: $(i)$ quantile binning ($9$ bins), $(ii)$ quantile binning with $10\%$ random reassignment ($6$ bins), $(iii)$ probabilistic sampling with weak target dependence ($10$ categories), $(iv)$ KMeans clustering on a 2D subspace ($4$ clusters), and $(v)$ quantile binning with $20\%$ random flips ($5$ bins). All categorical features are explicitly cast to discrete types. The original continuous features used to construct categorical variables are removed. 
The final dataset contains 15 predictors (10 continuous, 5 categorical) and one binary target.

\subsection{Probe Examples}\label{app:probe-example}

In this subsection, we provide example contamination probes for the \texttt{adult} dataset. Since the \texttt{real}, \texttt{like}, and \texttt{swapped} variants cannot be distinguished when inspecting an individual probe in isolation, we include only examples from the \texttt{real} variant (\Cref{tab:probeexample-real}) and the \texttt{obfuscated} variant (\Cref{tab:probeexample-obfuscated}).

\begin{table*}[htbp]
    \centering
    \caption{\textbf{Example probe of the \texttt{completion} and \texttt{existence} tasks for the \texttt{adult} dataset (\texttt{real} variant, coverage$=20\%$.} The upper block reports the probe for the completion task, while the lower block reports the probe for the existence task.}
    \vspace{0.5em}
    \label{tab:probeexample-real}

    \hrule height 1pt
    \vspace{0.5em}

    \textbf{COMPLETION}
    \vspace{0.3em}

\begin{lstlisting}
Complete the masked columns in this row with '?' marks. Choose the correct values.

ROW: age=?, workclass=Private, fnlwgt=198654, education=Prof-school, educational-num=15, marital-status=Married-civ-spouse, occupation=?, relationship=Husband, race=Asian-Pac-Islander, gender=Male, capital-gain=0, capital-loss=?, hours-per-week=67, native-country=India, income=>50K

OPTIONS:
1. occupation=Adm-clerical, capital-loss=1669, age=82
2. occupation=Farming-fishing, capital-loss=2282, age=84
3. occupation=Prof-specialty, capital-loss=2415, age=39
4. occupation=Farming-fishing, capital-loss=1421, age=37
5. occupation=Farming-fishing, capital-loss=1411, age=62
\end{lstlisting}

    \vspace{0.2em}
    \hrule height 0.6pt
    \vspace{0.5em}

    \textbf{EXISTENCE}
    \vspace{0.3em}

\begin{lstlisting}
Which of the following complete rows exists in the dataset? Choose the index only.

OPTIONS:
1. age=81, workclass=Federal-gov, fnlwgt=424340, education=Bachelors, educational-num=8, marital-status=Widowed, occupation=Prof-specialty, relationship=Not-in-family, race=White, gender=Female, capital-gain=0, capital-loss=1668, hours-per-week=3, native-country=Hungary, income<=50K
2. age=81, workclass=Self-emp-not-inc, fnlwgt=123959, education=Bachelors, educational-num=13, marital-status=Widowed, occupation=Prof-specialty, relationship=Not-in-family, race=White, gender=Female, capital-gain=0, capital-loss=1668, hours-per-week=3, native-country=Hungary, income<=50K
3. age=81, workclass=State-gov, fnlwgt=174778, education=Bachelors, educational-num=8, marital-status=Widowed, occupation=Prof-specialty, relationship=Not-in-family, race=White, gender=Female, capital-gain=0, capital-loss=1668, hours-per-week=3, native-country=Hungary, income<=50K
4. age=81, workclass=Federal-gov, fnlwgt=258761, education=Bachelors, educational-num=15, marital-status=Widowed, occupation=Prof-specialty, relationship=Not-in-family, race=White, gender=Female, capital-gain=0, capital-loss=1668, hours-per-week=3, native-country=Hungary, income<=50K
5. age=81, workclass=Local-gov, fnlwgt=353281, education=Bachelors, educational-num=16, marital-status=Widowed, occupation=Prof-specialty, relationship=Not-in-family, race=White, gender=Female, capital-gain=0, capital-loss=1668, hours-per-week=3, native-country=Hungary, income<=50K
\end{lstlisting}

    \vspace{0.2em}
    \hrule height 1pt
\end{table*}

\clearpage

\begin{table*}[htbp]
    \centering
    \caption{\textbf{Example probe of the \texttt{completion} and \texttt{existence} tasks for the \texttt{adult} dataset (\texttt{obfuscated} variant, coverage$=20\%$.} The upper block reports the probe for the completion task, while the lower block reports the probe for the existence task.}
    \vspace{0.5em}
    \label{tab:probeexample-obfuscated}

    \hrule height 1pt
    \vspace{0.5em}

    \textbf{COMPLETION}
    \vspace{0.3em}

\begin{lstlisting}
Complete the masked columns in this row with '?' marks. Choose the correct values.

ROW: f1=?, f2=c2, f3=198654, f4=c14, f5=15, f6=c2, f7=?, f8=c0, f9=c1, f10=c1, f11=0, f12=?, f13=67, f14=c18, target=1

OPTIONS:
1. f7=c8, f12=1669, f1=82
2. f7=c9, f12=2282, f1=84
3. f7=c1, f12=2415, f1=39
4. f7=c3, f12=1421, f1=37
5. f7=c3, f12=1411, f1=62
\end{lstlisting}

    \vspace{0.2em}
    \hrule height 0.6pt
    \vspace{0.5em}

    \textbf{EXISTENCE}
    \vspace{0.3em}

\begin{lstlisting}
Which of the following complete rows exists in the dataset? Choose the index only.

OPTIONS:
1. f1=81, f2=c000, f3=424340, f4=c009, f5=8, f6=c006, f7=c009, f8=c001, f9=c004, f10=c000, f11=0, f12=1668, f13=3, f14=c017, target=0
2. f1=81, f2=c004, f3=123959, f4=c009, f5=13, f6=c006, f7=c009, f8=c001, f9=c004, f10=c000, f11=0, f12=1668, f13=3, f14=c017, target=0
3. f1=81, f2=c005, f3=174778, f4=c009, f5=8, f6=c006, f7=c009, f8=c001, f9=c004, f10=c000, f11=0, f12=1668, f13=3, f14=c017, target=0
4. f1=81, f2=c000, f3=258761, f4=c009, f5=15, f6=c006, f7=c009, f8=c001, f9=c004, f10=c000, f11=0, f12=1668, f13=3, f14=c017, target=0
5. f1=81, f2=c001, f3=353281, f4=c009, f5=16, f6=c006, f7=c009, f8=c001, f9=c004, f10=c000, f11=0, f12=1668, f13=3, f14=c017, target=0
\end{lstlisting}

    \vspace{0.2em}
    \hrule height 1pt
\end{table*}

\section{LLM Interrogation Protocol}\label{app:interrogation-protocol}

Each probe is converted into a standardized prompt template. The model is told that it will receive a multiple-choice question about a tabular dataset, that the options are numbered from $1$ to $5$, and that it should first reason step by step and then produce a final answer consisting of a single integer. 

To bypass the limitations of constrained decoding while supporting intermediate reasoning \citep{constrainedgeneration-problem, CoT}, models are prompted to generate free-form text before outputting their final integer choice within a programmatically parsed tag \citep{constrainedgeneration-hybrid}. Therefore, the final answer must be enclosed within explicit delimiters to allow reliable post-processing. This yields a robust, machine-parseable evaluation pipeline while still allowing the model to produce an unconstrained rationale.

Tables~\ref{tab:promptexample-real} and~\ref{tab:promptexample-obfuscated} illustrate the interrogation protocol through a prompt example for the \texttt{real} and \texttt{obfuscated} variant, respectively. Notice that in the \texttt{real} variant (the same holds for \texttt{like} and \texttt{swapped}), we explicitly specify the dataset name, since we want the model to leverage prior knowledge about it, if it has any. In contrast, in the \texttt{obfuscated} variant, we name it as ``a tabular dataset", describing only the number and type of features.

\begin{table}[htbp]
    \centering
    \caption{\textbf{Example prompt for the completion task on the \texttt{iris} dataset (\texttt{real} variant, coverage=$20\%$).} The prompt specifies the response format, provides a row with one masked feature, and asks the model to identify the correct value from five multiple-choice candidates.}
    \vspace{0.5em}
    \label{tab:promptexample-real}
    
    \hrule height 1pt 
    \vspace{0.5em}
    
\begin{lstlisting}
You will be given a MULTIPLE-CHOICE question about the Iris dataset.
Each question is MULTIPLE-CHOICE with options numbered from 1 to 5.

IMPORTANT INSTRUCTIONS:
1. First, reason step-by-step about the question and each option. Think through which option is most plausible.
2. After your reasoning, write your final answer as a single integer between 1 and 5 (the number of the chosen option).
3. You MUST wrap your final answer between these exact tags: [FINAL_ANSWER_start] <number> [FINAL_ANSWER_end]
   Example: [FINAL_ANSWER_start] 3 [FINAL_ANSWER_end]
4. If you are uncertain, still choose the single most plausible option.

QUESTION:
Complete the masked columns in this row with '?' marks. Choose the correct values.
ROW: sepal_length=?, sepal_width=3.3, petal_length=1.7, petal_width=0.5, species=Iris-setosa

OPTIONS:
1: sepal_length=6.1
2: sepal_length=5.1
3: sepal_length=5.0
4: sepal_length=6.8
5: sepal_length=7.0

First reason about the question, then provide your final answer between the tags.
\end{lstlisting}

    \vspace{0.2em}
    \hrule height 1pt 
\end{table}

\begin{table}[htbp]
    \centering
    \caption{\textbf{Example prompt for the completion task on the \texttt{iris} dataset (\texttt{obfuscated} variant, coverage=$20\%$).} The prompt specifies anonymized information about the dataset, the response format, provides a row with one masked feature, and asks the model to identify the correct value from five multiple-choice candidates.}
    \vspace{0.5em}
    \label{tab:promptexample-obfuscated}
    
    \hrule height 1pt 
    \vspace{0.5em}
    
\begin{lstlisting}
You will be given a MULTIPLE-CHOICE question about a tabular dataset. This dataset contains 4 features in total. They are all numerical features:f1, f2, f3, f4. The target variable is categorical with 3 classes.

Each question is MULTIPLE-CHOICE with options numbered from 1 to 5.

IMPORTANT INSTRUCTIONS:
[...]

QUESTION:
Complete the masked columns in this row with '?' marks. Choose the correct values.\nROW: f1=?, f2=3.3, f3=1.7, f4=0.5, target=c0

OPTIONS:
1: f1=6.1
2: f1=5.1
3: f1=5.0
4: f1=6.8
5: f1=7.0

First reason about the question, then provide your final answer between the tags.
\end{lstlisting}

    \vspace{0.2em}
    \hrule height 1pt 
\end{table}

\newpage

\section{Probability Distribution of Options in Probes}\label{app: baseline_prob_build}

\subsection{Deterministic Marginal Log-Likelihood Baseline}

The deterministic baseline scores each option using column-wise empirical marginals estimated from the original cleaned dataset. For each categorical column $c$, if a value $v$ appears $\mathrm{count}_c(v)$ times, the baseline estimates
\begin{equation*}
p(v \mid c) = \frac{\mathrm{count}_c(v)+1}{N_c + |\mathcal{V}_c| + 1},
\end{equation*}
where $N_c$ is the number of non-missing observations in column $c$ and $|\mathcal{V}_c|$ is the number of distinct observed values. This corresponds to Laplace smoothing. 
Similarly, for the numerical columns $c$, we apply the \textit{binning} approach. Specifically, we compute the number of bins using Scott's rule \citep{scott-rule}, thus obtaining a set of intervals $\{I_1,\dots,I_B\}$. \\
Next, the baseline estimate 
\begin{equation*}
p(v \mid c) = p(\text{bin}(v)\mid c)=\frac{\mathrm{count}_{\text{bin}}(v)+1}{N_c+B}
\end{equation*}
where $B$ is the number of bins and $\mathrm{count}_{\text{bin}}(v)$ returns the number of observations which are within the bin of $v$.
Finally, given an option $o$, represented as a set of column-value assignments, the score is defined as
\begin{equation*}
S(o) = \sum_{(c,v)\in o} \log p(v \mid c).
\end{equation*}
Unseen values receive a large negative default score. 

The deterministic baseline selects the option with the maximal score:
\begin{equation*}
\hat{y} = \arg\max_o S(o),
\end{equation*}
with random tie-breaking when necessary.
For both completion and existence probes, $o$ includes only the masked columns. In the \texttt{obfuscated} variant, options are first translated back into the original feature/value space via stored metadata, and only then scored against the real-data marginals.
In contrast, in \texttt{swapped} and \texttt{like} variants, we compute the score directly against the real-data marginals. This makes the baseline comparable across variants.

\subsection{Stochastic Marginal Log-Likelihood Baseline}

The stochastic baseline uses the same score $S(o)$ as above, but converts scores into a softmax distribution:
\begin{equation*}
P(\hat{y}=o_i) = \frac{\exp(S(o_i)-\max_j S(o_j))}{\sum_j \exp(S(o_j)-\max_k S(o_k))}.
\end{equation*}
An option is then sampled according to this distribution. This baseline captures what can be achieved by exploiting marginal column statistics without making a purely greedy decision.

\newpage

\section{Additional Results}\label{app: additional-results}

\begin{figure}[htpb]
    \centering
    
    \begin{subfigure}[b]{1\textwidth}
        \centering
        \includegraphics[width=\linewidth]{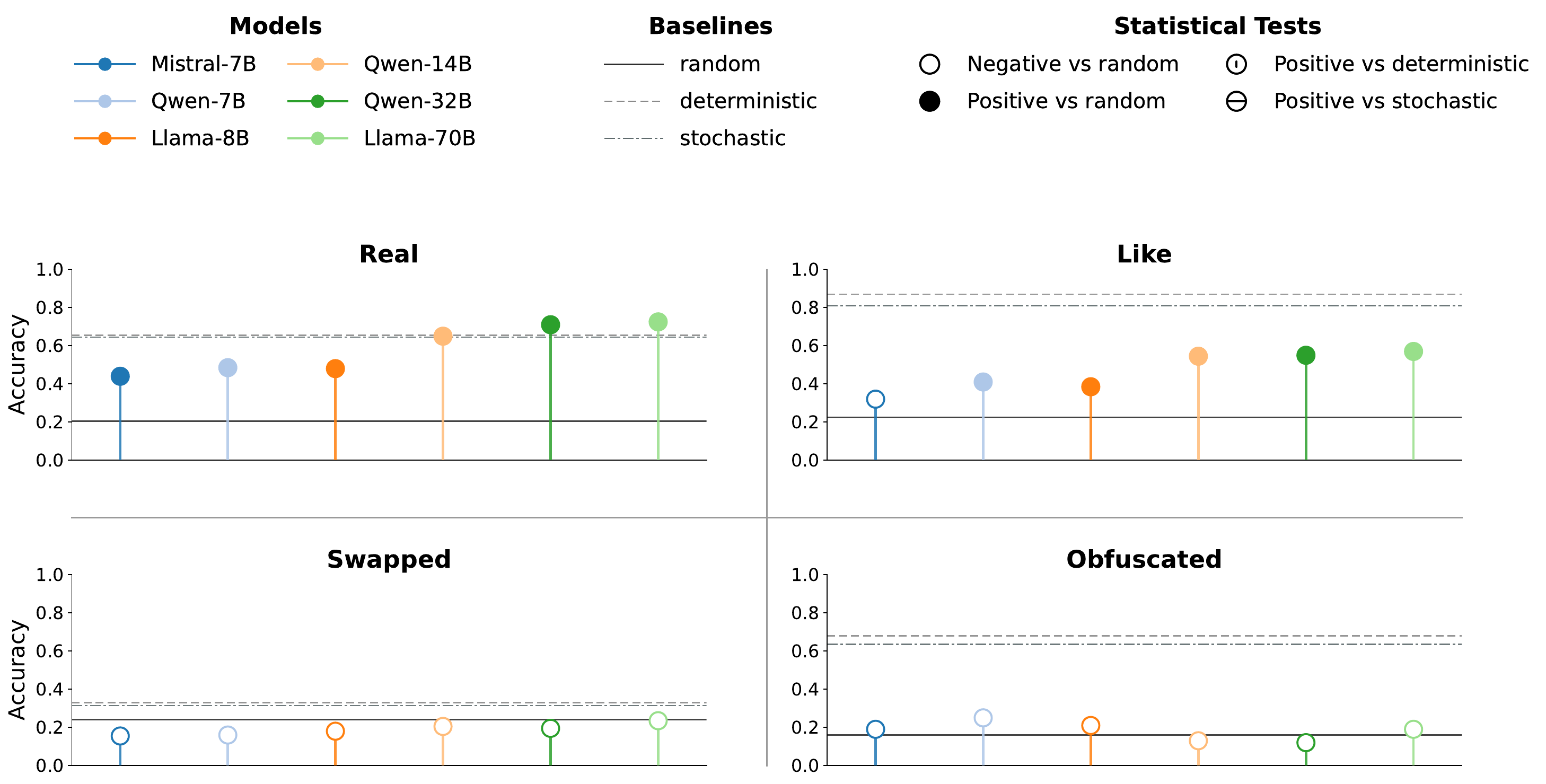}
        \caption{Completion of \texttt{adult}}
        \label{fig:completion-adult}
    \end{subfigure}
    
    \vspace{1.5em} 
    
    \begin{subfigure}[b]{1\textwidth}
        \centering
        \includegraphics[width=\linewidth]{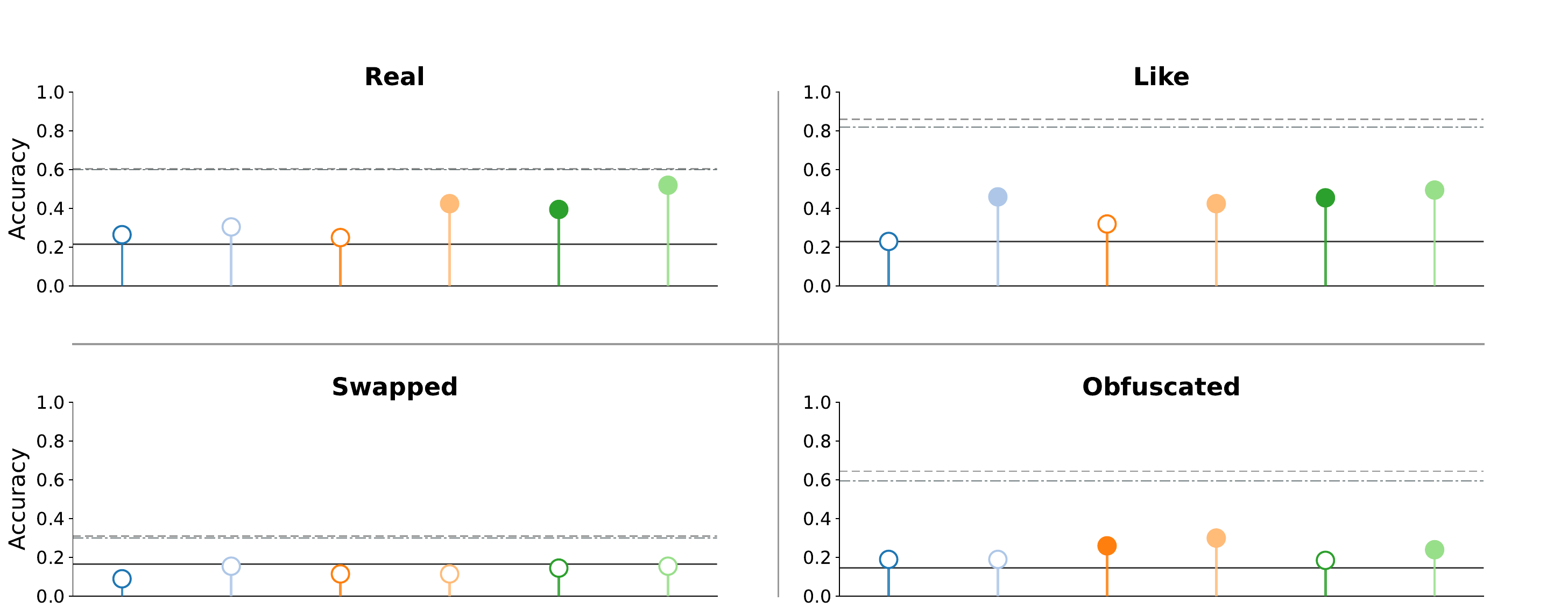}
        \caption{Existence of \texttt{adult}}
        \label{fig:existence-adult}
    \end{subfigure}
    
    \caption{\textbf{Task performance for \texttt{adult} across dataset variants and models.} We represent baselines as horizontal lines, and statistical test results using filled dots for detected contamination (positive against random), and horizontal/vertical bars on dots for superiority against statistical baselines.}
    \label{fig:combined-adult}
\end{figure}

\begin{figure}[htpb]
    \centering
    
    \begin{subfigure}[b]{1\textwidth}
        \centering
        \includegraphics[width=\linewidth]{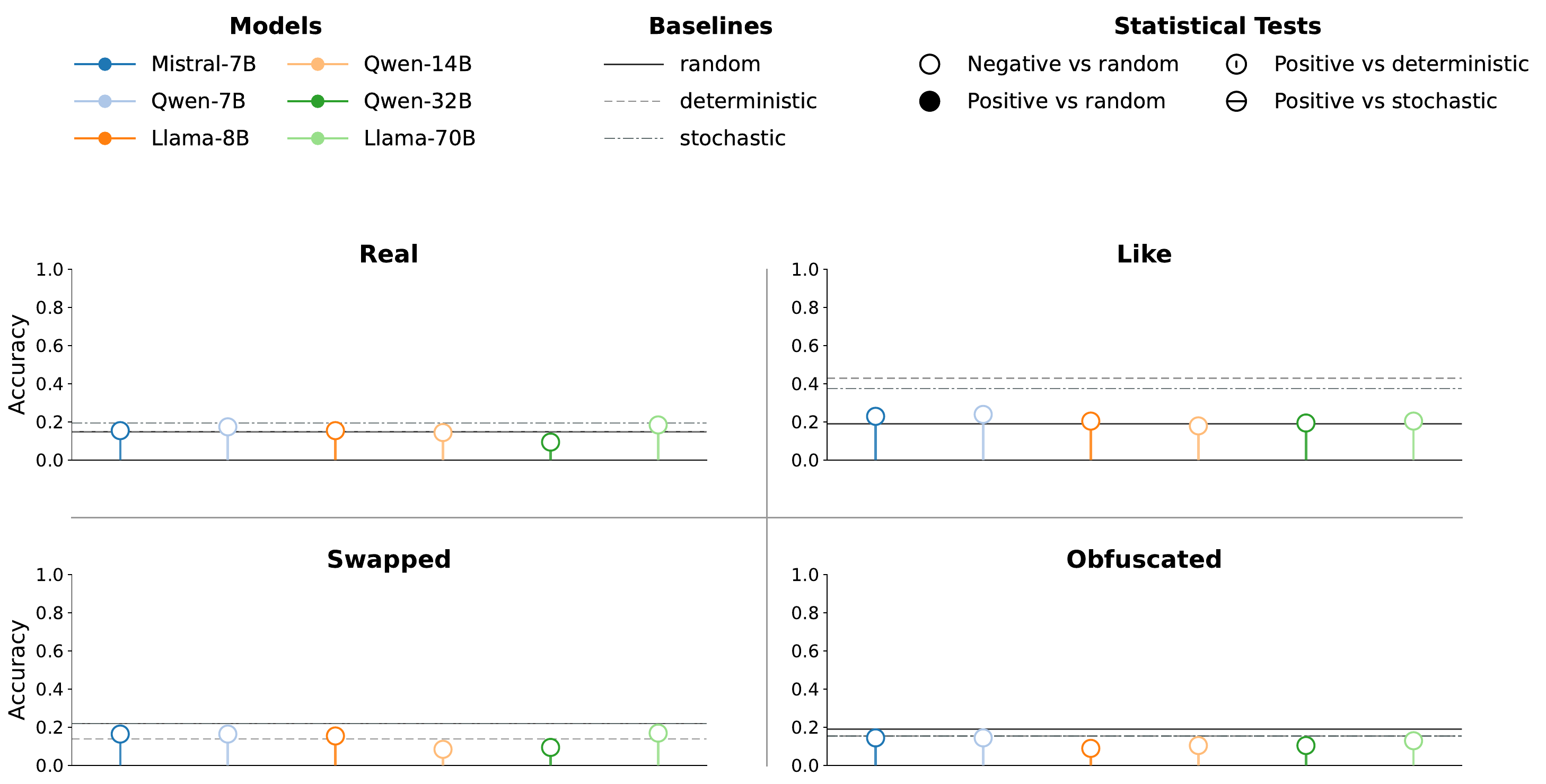}
        \caption{Completion of \texttt{synthetic}}
        \label{fig:completion-synthetic}
    \end{subfigure}
    
    \vspace{1.5em} 
    
    \begin{subfigure}[b]{1\textwidth}
        \centering
        \includegraphics[width=\linewidth]{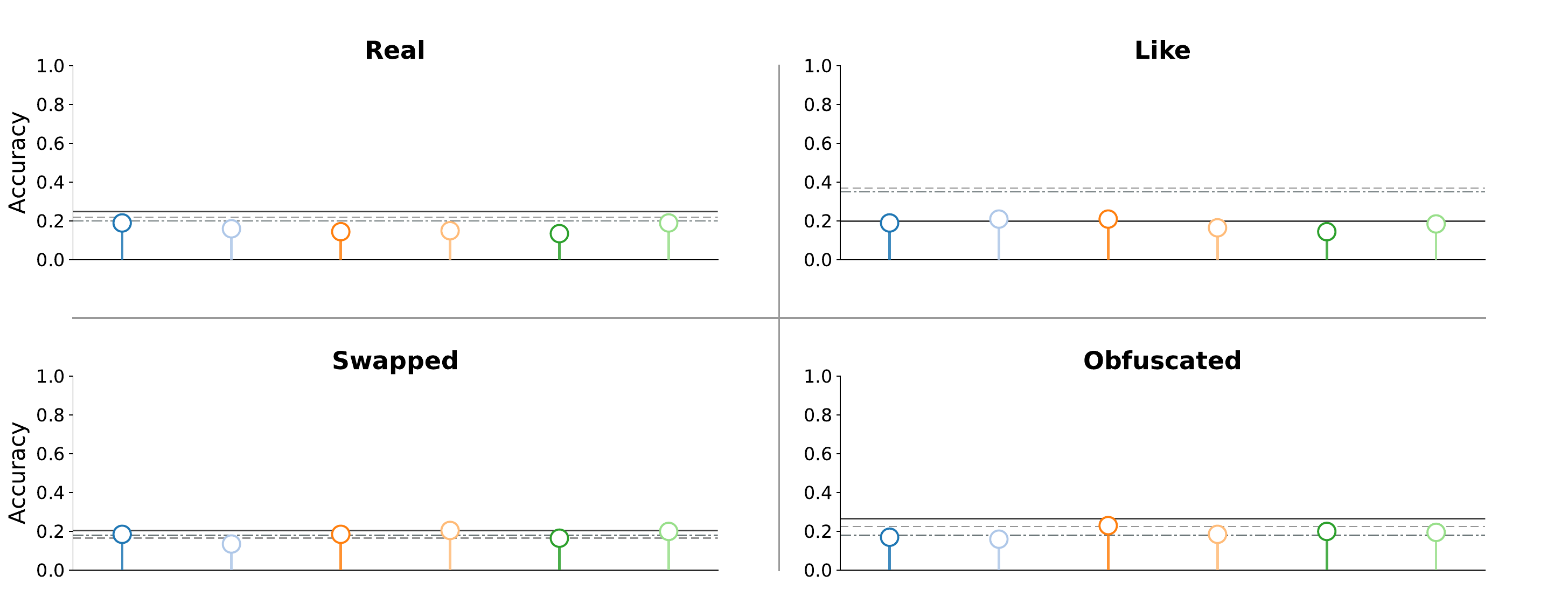}
        \caption{Existence of \texttt{synthetic}}
        \label{fig:existence-synthetic}
    \end{subfigure}
    
    \caption{\textbf{Task performance for \texttt{synthetic} across dataset variants and models.} We represent baselines as horizontal lines, and statistical test results using filled dots for detected contamination (positive against random), and horizontal/vertical bars on dots for superiority against statistical baselines.}
    \label{fig:combined-synthetic}
\end{figure}

\begin{figure}[htpb]
    \centering
    
    \begin{subfigure}[b]{1\textwidth}
        \centering
        \includegraphics[width=\linewidth]{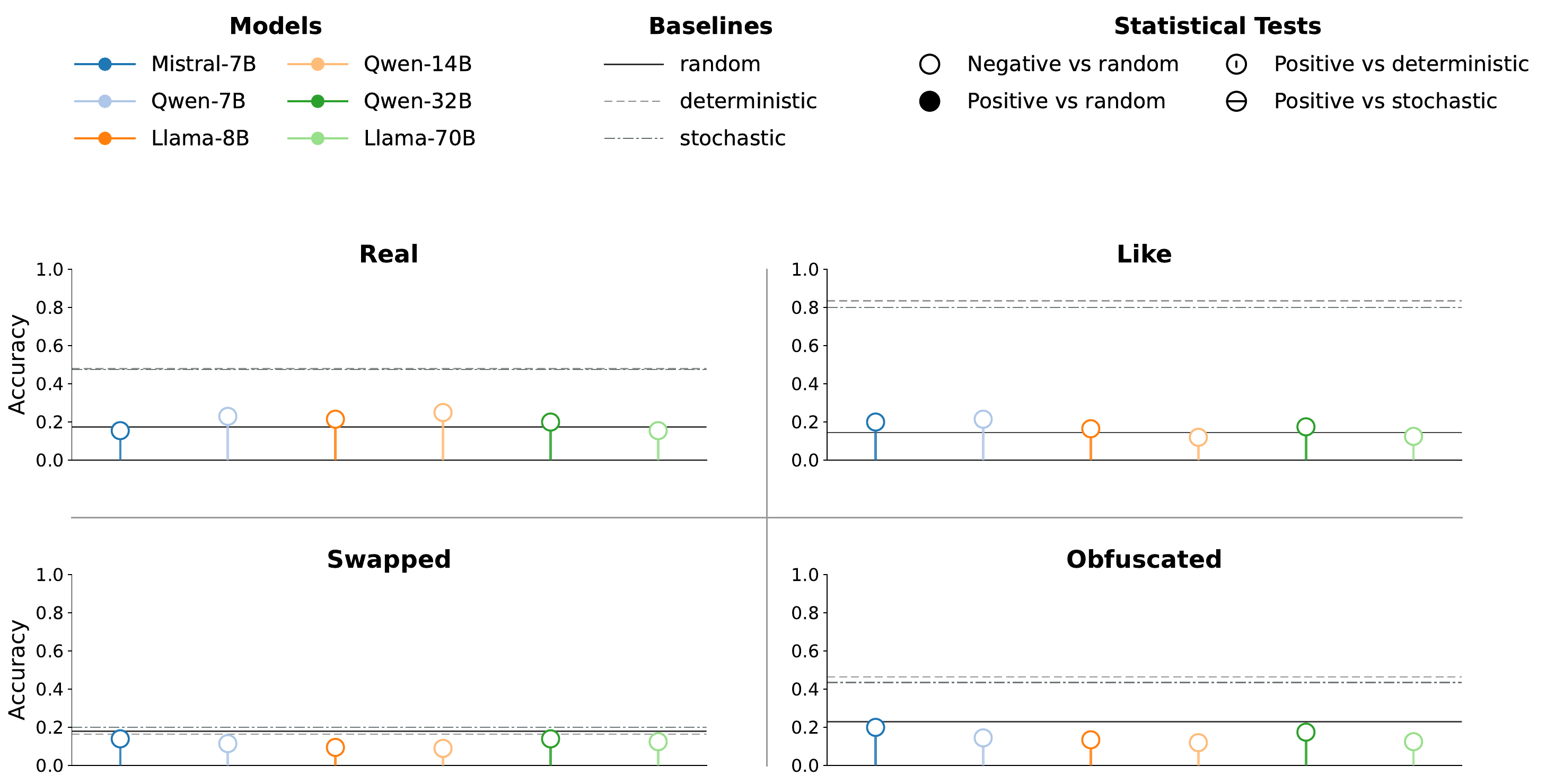}
        \caption{Completion of \texttt{mushroom}}
        \label{fig:completion-mushroom-app}
    \end{subfigure}
    
    \vspace{1.5em} 
    
    \begin{subfigure}[b]{1\textwidth}
        \centering
        \includegraphics[width=\linewidth]{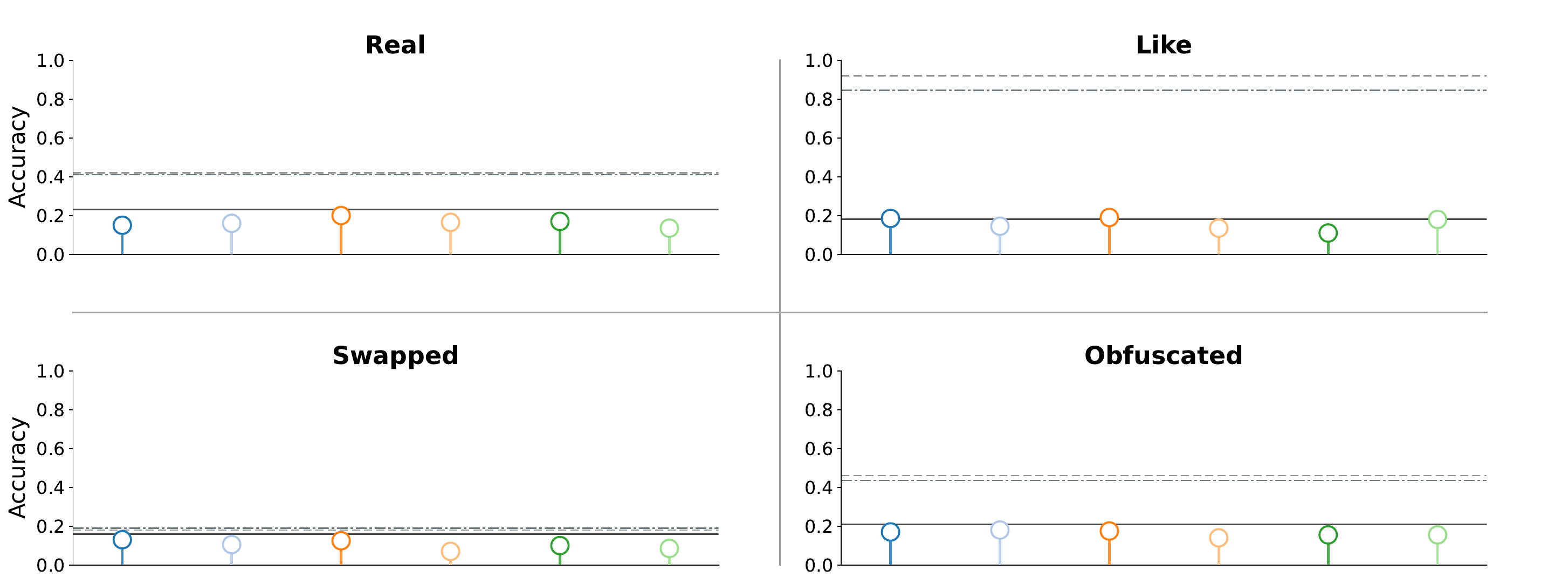}
        \caption{Existence of \texttt{mushroom}}
        \label{fig:existence-mushroom}
    \end{subfigure}
    
    \caption{\textbf{Task performance for \texttt{mushroom} across dataset variants and models.} We represent baselines as horizontal lines, and statistical test results using filled dots for detected contamination (positive against random), and horizontal/vertical bars on dots for superiority against statistical baselines.}
    \label{fig:combined-mushroom}
\end{figure}

\begin{figure}[htpb]
    \centering
    
    \begin{subfigure}[b]{1\textwidth}
        \centering
        \includegraphics[width=\linewidth]{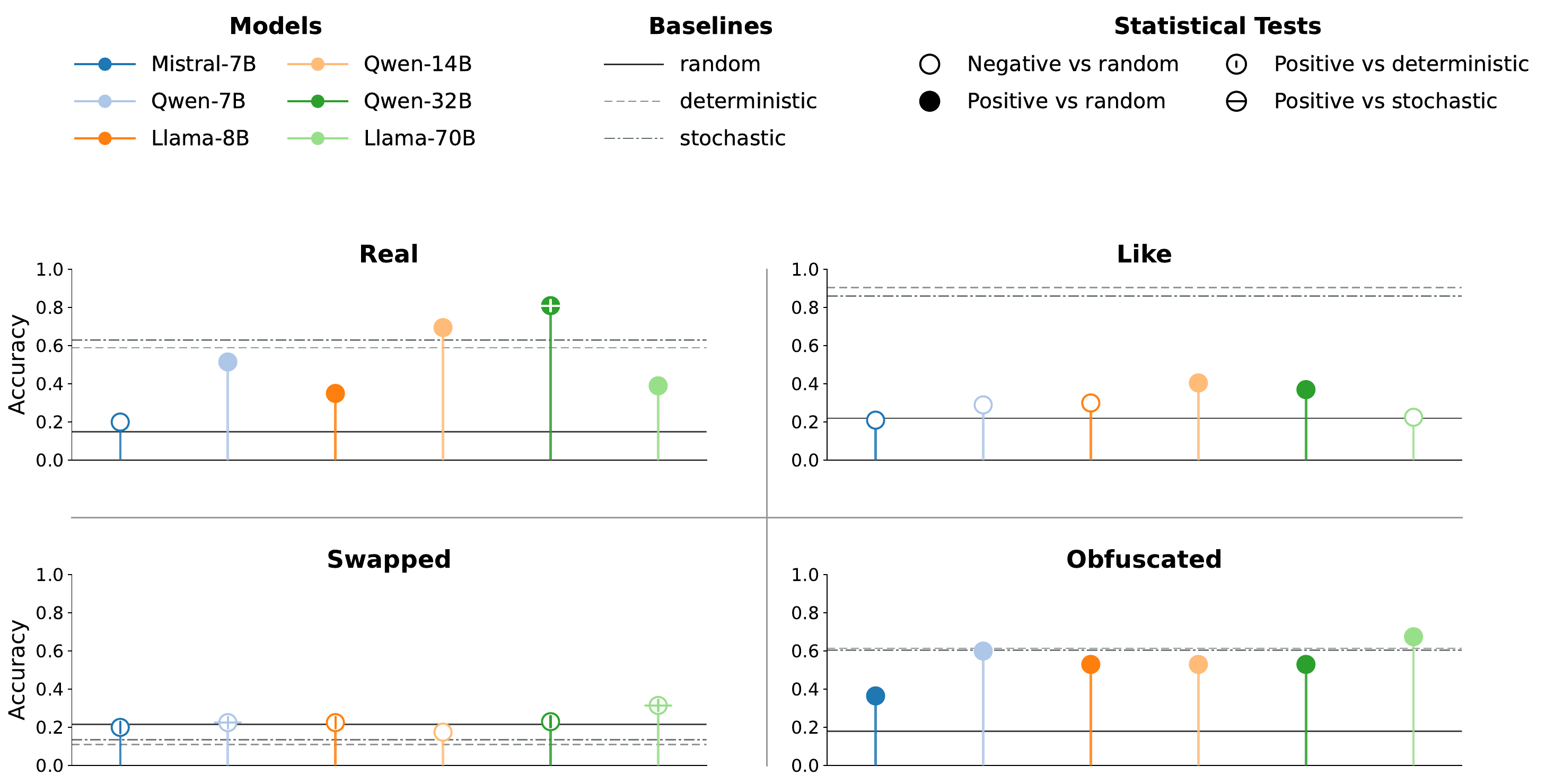}
        \caption{Completion of \texttt{credit}}
        \label{fig:completion-credit}
    \end{subfigure}
    
    \vspace{1.5em} 
    
    \begin{subfigure}[b]{1\textwidth}
        \centering
        \includegraphics[width=\linewidth]{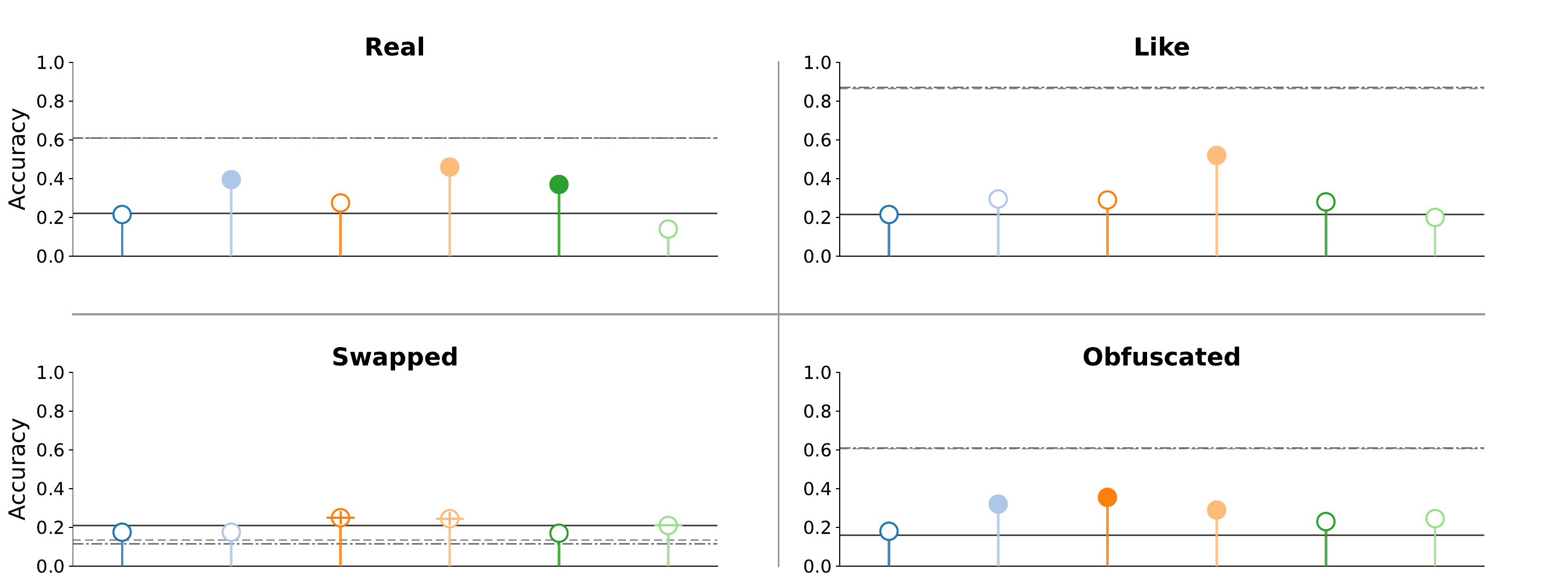}
        \caption{Existence of \texttt{credit}}
        \label{fig:existence-credit}
    \end{subfigure}
    
    \caption{\textbf{Task performance for \texttt{credit} across dataset variants and models.} We represent baselines as horizontal lines, and statistical test results using filled dots for detected contamination (positive against random), and horizontal/vertical bars on dots for superiority against statistical baselines.}
    \label{fig:combined-credit}
\end{figure}

In this section, we report additional figures, i.e., \Cref{fig:combined-adult,fig:combined-synthetic,fig:combined-mushroom,fig:combined-credit}. In \Cref{app:ablation}, we conduct an ablation on the hyperparameters of the framework, i.e., the temperature of the model and the level of coverage. Lastly, in \Cref{app:raw-results} we report raw results for all settings.

\clearpage

\subsection{Ablation Study}\label{app:ablation}

The goal of this section is to analyze how the outcomes of our framework change as two main hyperparameters vary: the LLM's sampling temperature and the percentage of columns masked or perturbed during probe construction, which we refer to as \emph{coverage}.

\subsubsection{Impact of Temperature}
We first study the effect of the decoding temperature. As shown in the right panels of \Cref{fig:real-transition}, changing the temperature has little effect on contamination detection on the real dataset variant. In most model--dataset pairs, the test outcome remains unchanged when moving from deterministic decoding to a more stochastic setting. This indicates that the observed results are not driven by sampling-induced response variability and suggests that the framework is robust to this factor.

\begin{figure}[htpb]
    \centering
    
    \begin{subfigure}[b]{1\textwidth}
        \centering
        \includegraphics[width=1\linewidth]{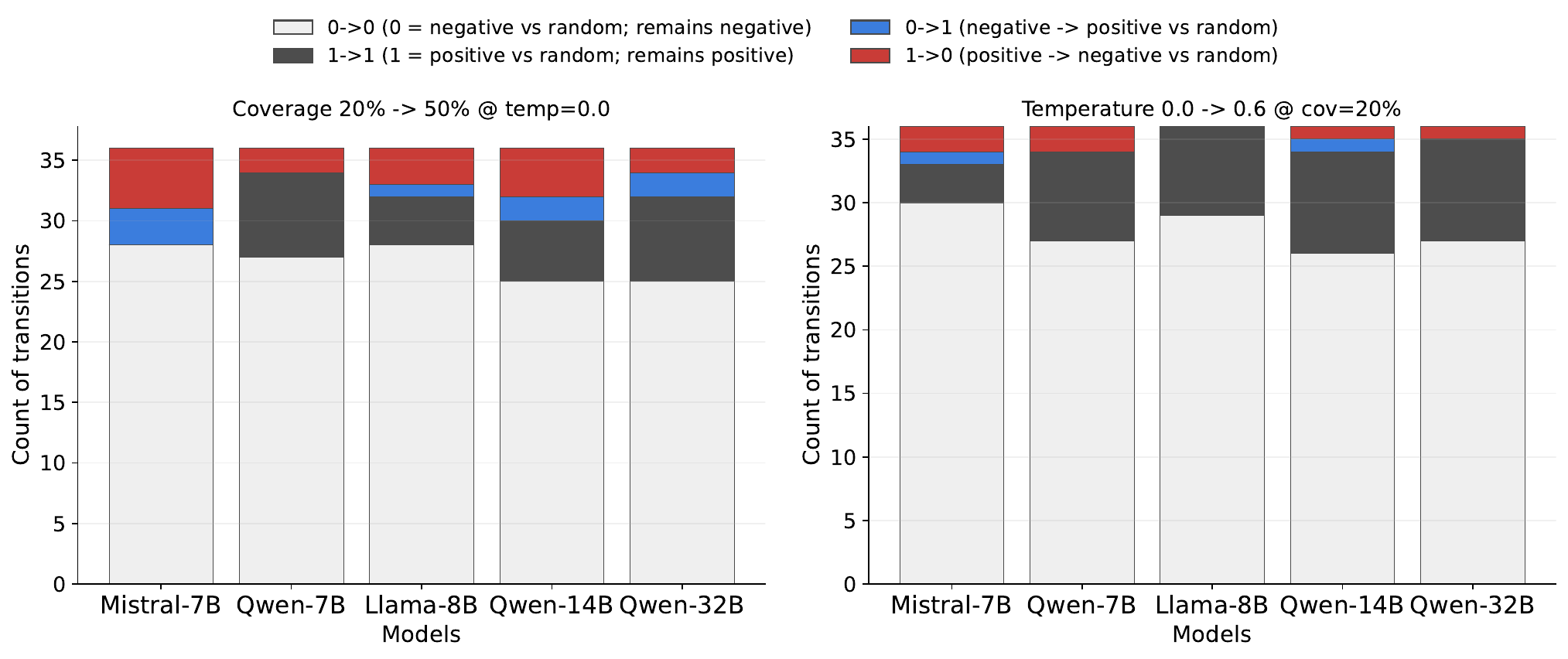}
        \caption{Completion}
        \label{fig:all-summary-completion}
    \end{subfigure}
    
    \vspace{1.5em} 
    
    \begin{subfigure}[b]{1\textwidth}
        \centering
        \includegraphics[width=1\linewidth]{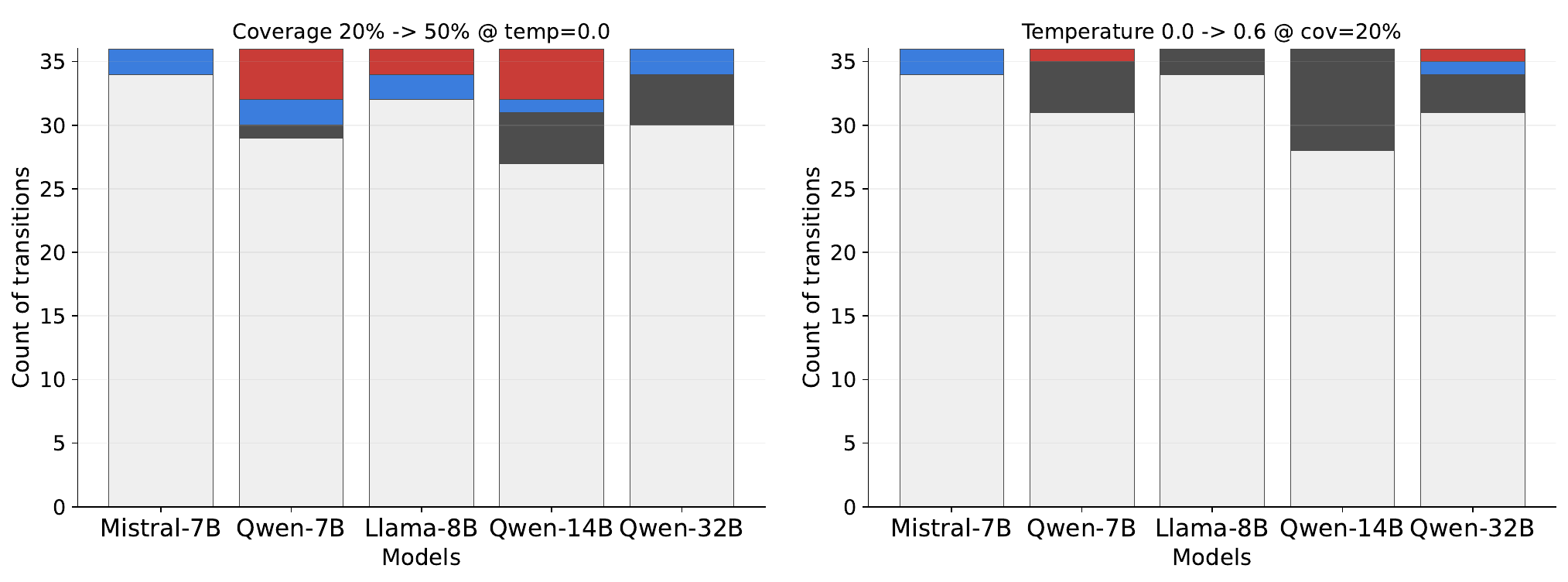}
        \caption{Existence}
        \label{fig:all-summary-existence}
    \end{subfigure}
    
    \caption{\textbf{Aggregate transition counts across all dataset variants for the completion and existence task}. For each model, the bars report how many contamination-test outcomes remain unchanged or switch sign when increasing coverage from 
    $20\%$ to $50\%$ at fixed temperature $t=0$ (left), or temperature from $0$ to $0.6$ at $20\%$ coverage (right). Colors follow the transition categories defined in the legend.}
    \label{fig:all-summary}
\end{figure}

The same pattern is confirmed by the aggregate counts reported in the right panels of \Cref{fig:all-summary}, which summarize transitions across all dataset variants for each model. Nearly all models exhibit only a small number of outcome changes, and in some cases, such as \textit{Llama-8B}, the effect is negligible. Overall, these results show that temperature is not a critical factor in the framework's behavior.

\subsubsection{Impact of Coverage}
We next analyze the effect of coverage, i.e., the fraction of columns that are masked or perturbed in the generated probes. Unlike temperature, coverage directly affects the task's difficulty and, therefore, performance on the real variant.

\begin{figure}[htpb]
    \centering
    
    \begin{subfigure}[b]{1\textwidth}
        \centering
        \includegraphics[width=1\linewidth]{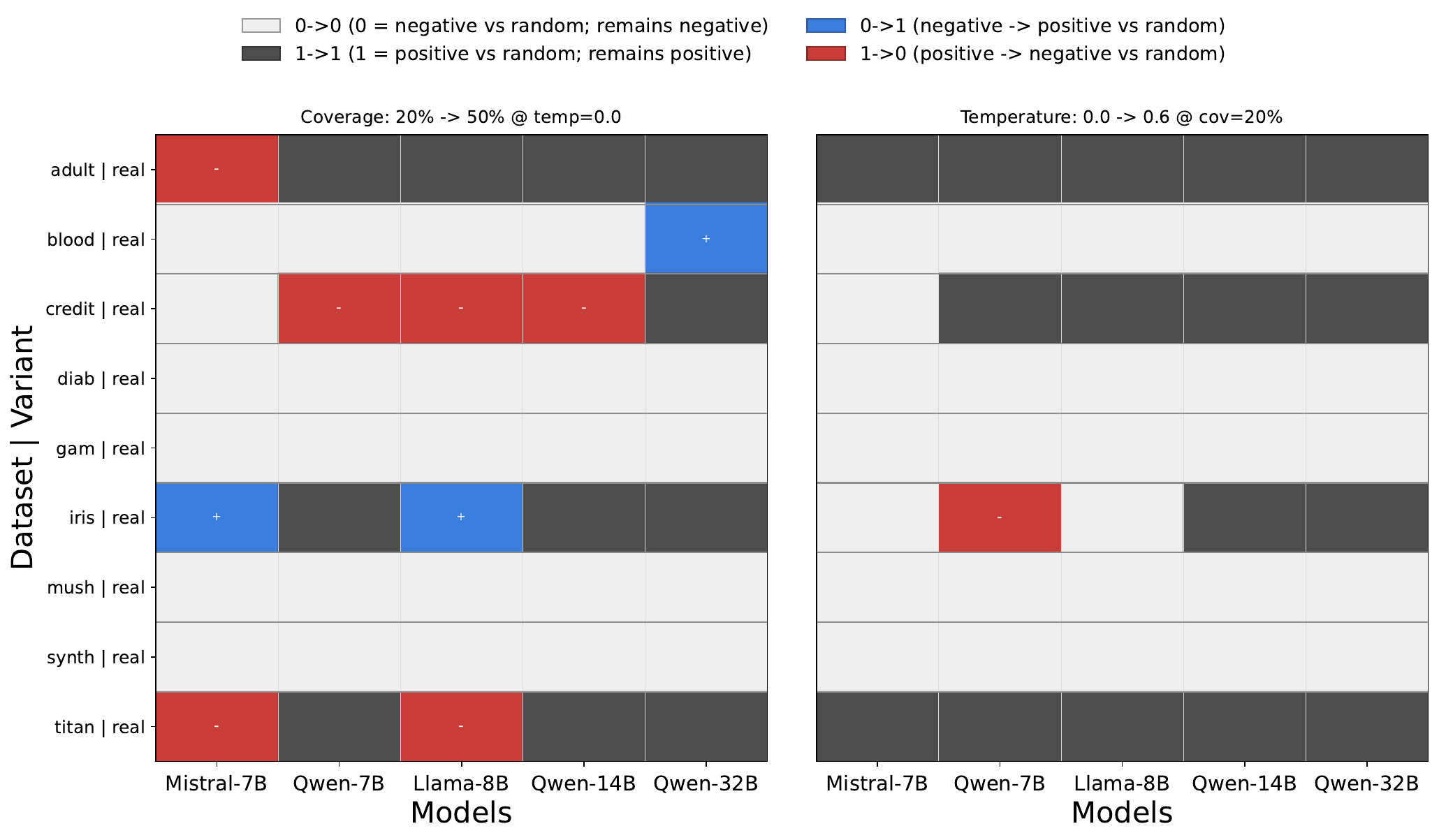}
        \caption{Completion}
        \label{fig:real-transition-completion}
    \end{subfigure}
    
    \vspace{1.5em} 
    
    \begin{subfigure}[b]{1\textwidth}
        \centering
        \includegraphics[width=1\linewidth]{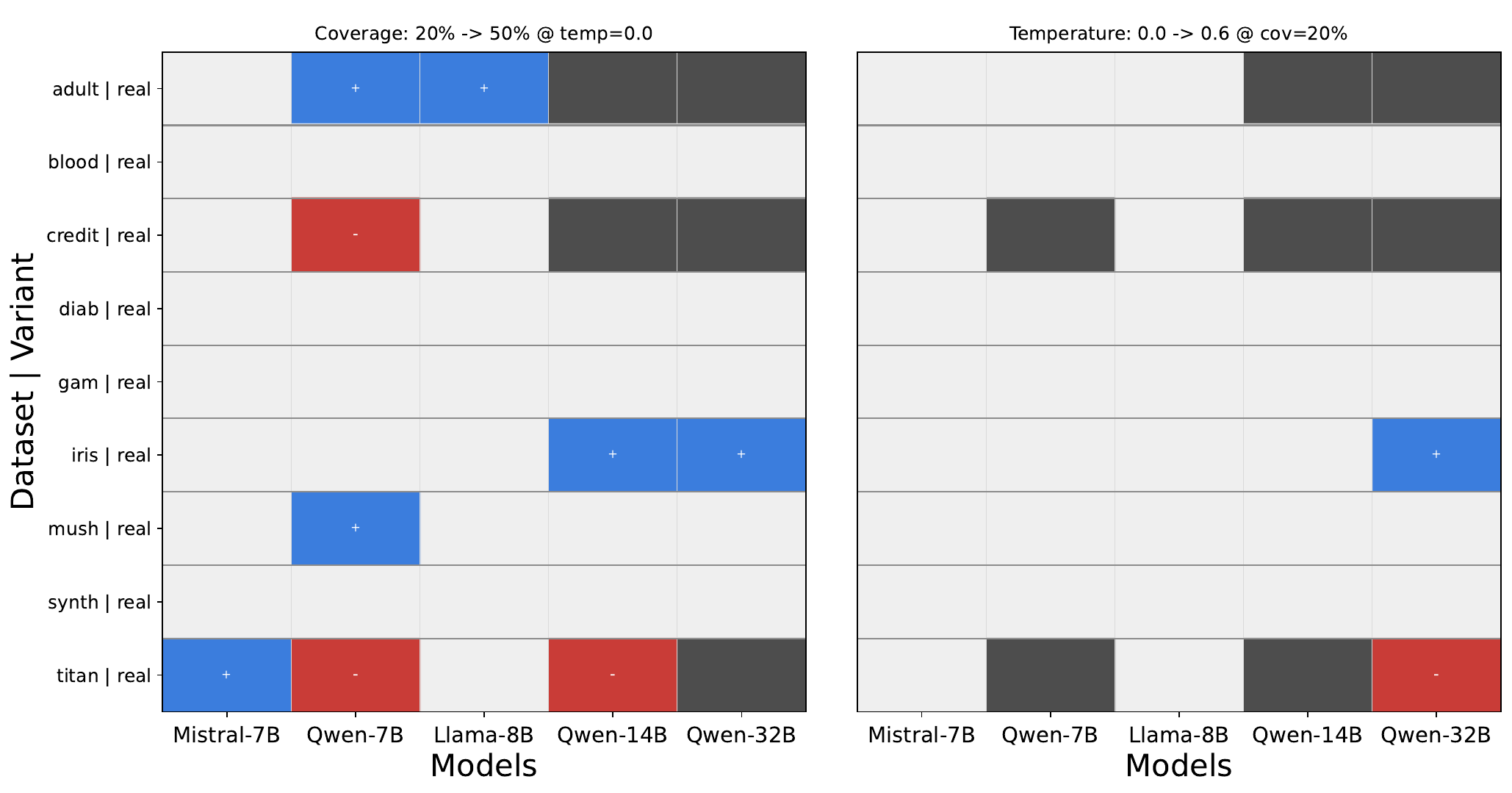}
        \caption{Existence}
        \label{fig:real-transition-existence}
    \end{subfigure}
    
    \caption{\textbf{Ablation results on the real dataset variant for the completion and existence task}. The figure shows transitions in contamination-test outcomes for each model--dataset pair when increasing coverage from $20\%$ to $50\%$ at fixed temperature $t=0$ (left), or temperature from $0$ to $0.6$ at $20\%$ coverage (right)
    Colors denote stable outcomes and positive/negative flips, as reported in the legend.}
    \label{fig:real-transition}
\end{figure}

\begin{figure}[htpb]
    \centering
    
    \begin{subfigure}[b]{1\textwidth}
        \centering
        \includegraphics[width=1\linewidth]{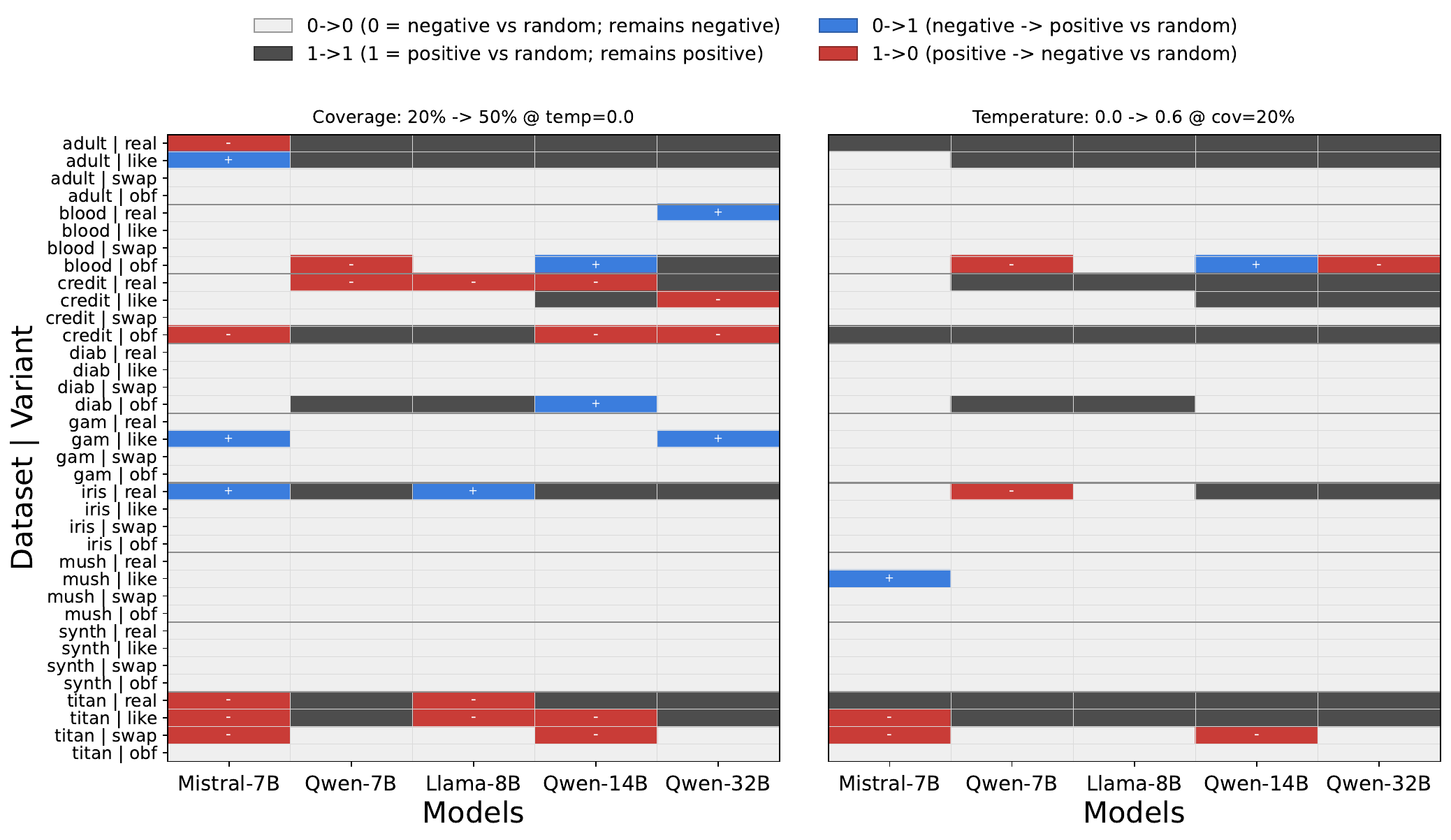}
        \caption{Completion}
        \label{fig:all-transition-completion}
    \end{subfigure}
    
    \vspace{1.5em} 
    
    \begin{subfigure}[b]{1\textwidth}
        \centering
        \includegraphics[width=1\linewidth]{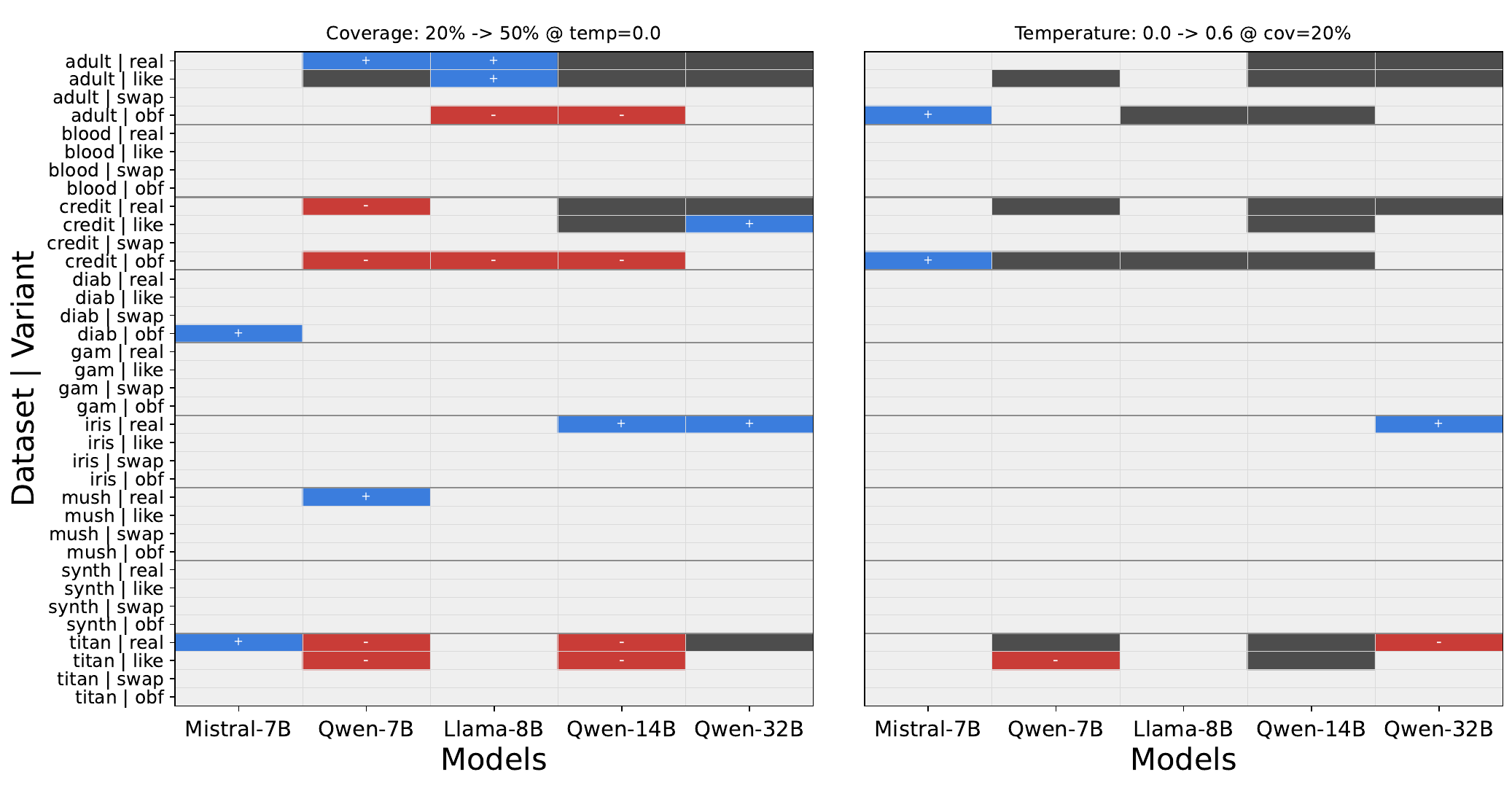}
        \caption{Existence}
        \label{fig:all-transition-existence}
    \end{subfigure}
    
    \caption{\textbf{Ablation results on dataset variants for the completion and existence task}. The figure shows transitions in contamination-test outcomes for each model--dataset pair when increasing coverage from $20\%$ to $50\%$ at fixed temperature $t=0$ (left), or temperature from $0$ to $0.6$ at $20\%$ coverage (right). Colors denote stable outcomes and positive/negative flips, as reported in the legend.}
    \label{fig:all-transition}
\end{figure}

For the completion task, the left panels of \Cref{fig:real-transition-completion,fig:all-transition-completion} show that most transitions occur from contamination detected to no contamination detected as coverage increases. This behavior is expected: masking a larger fraction of columns makes the completion problem harder, reducing the information available to recover the correct answer.

For the existence task, the trend is different. As shown in the left panels of \Cref{fig:real-transition-existence,fig:all-transition-existence}, increasing coverage often makes the task easier rather than harder, with several transitions in the opposite direction. Adding more information to the distractor options makes the true row easier to identify. This suggests that LLMs can exploit prior knowledge effectively in this setting and, somewhat surprisingly, may benefit from stronger perturbations applied to the negatives.

\subsubsection{Summary}
Overall, this ablation study shows that temperature has a limited influence on the conclusions of our framework, whereas coverage plays a substantial role, and a more principled treatment of this hyperparameter would further strengthen the framework. One direction is to run probes at multiple coverage levels and aggregate the resulting outcomes, so that the contamination verdict depends on a model's behavior across the entire coverage range rather than on a single, fixed choice. We leave this multi-coverage analysis, along with the design of perturbation strategies that reduce the sensitivity of performance to coverage, to future work.


\newpage

\subsection{Detailed Contamination Results}\label{app:raw-results}

In this section, we report the complete raw results across all models and datasets used throughout the paper, including those summarized in the main text and discussed in the previous section. Specifically, \Cref{tab:contamination_summary_t0_cov20} presents results with temperature $= 0$ and coverage $= 20\%$, \Cref{tab:contamination_summary_t0.6_cov20} reports results with temperature $= 0.6$ and coverage $= 20\%$, and \Cref{tab:contamination_summary_t0_cov50} shows results with temperature $= 0$ and coverage $= 50\%$.

\setlength{\tabcolsep}{4pt}
\begin{xltabular}{0.9\textwidth}{l l l c c c c c c}
\caption{\textbf{Summary of contamination results across datasets and models (temperature = 0.0, coverage = 20\%).} $\mathcal{A}_{\mathrm{comp}}$ denotes completion accuracy, and $\mathcal{A}_{\mathrm{exist}}$ denotes existence accuracy. Bold values indicate performances that are statistically significant above the random baseline and have higher accuracy than the random baseline ($\alpha = 0.01$); $^{*}$ indicates significance against the deterministic baseline, the stochastic baseline, or both ($\alpha = 0.01$).}\label{tab:contamination_summary_t0_cov20} \\
\toprule
\multirow{2}{*}{\centering\textbf{Dataset}} & \multirow{2}{*}{\centering\textbf{Variant}} & \multirow{2}{*}{\centering\textbf{Metric}} & \multicolumn{6}{c}{\textbf{Model}} \\
\cmidrule(lr){4-9}
 &  &  & \textit{Mistral-7B} & \textit{Qwen-7B} & \textit{Llama-8B} & \textit{Qwen-14B} & \textit{Qwen-32B} & \textit{Llama-70B} \\
\midrule
\endfirsthead
\caption[]{Summary of contamination results across datasets and models (continued)} \\
\toprule
\multirow{2}{*}{\centering\textbf{Dataset}} & \multirow{2}{*}{\centering\textbf{Variant}} & \multirow{2}{*}{\centering\textbf{Metric}} & \multicolumn{6}{c}{\textbf{Model}} \\
\cmidrule(lr){4-9}
 &  &  & \textit{Mistral-7B} & \textit{Qwen-7B} & \textit{Llama-8B} & \textit{Qwen-14B} & \textit{Qwen-32B} & \textit{Llama-70B} \\
\midrule
\endhead
\midrule
\multicolumn{9}{r}{\textit{Continued on next page}} \\
\midrule
\endfoot
\bottomrule
\endlastfoot
\multirow{8}{*}{\texttt{adult}} & \multirow{2}{*}{real} & \g $\mathcal{A}_{\mathrm{comp}}$ & \g $\mathbm{0.44}$ & \g $\mathbm{0.48}$ & \g $\mathbm{0.48}$ & \g $\mathbm{0.65}$ & \g $\mathbm{0.71}$ & \g $\mathbm{0.72}$ \\
 &  & $\mathcal{A}_{\mathrm{exist}}$ & $0.27$ & $0.30$ & $0.25$ & $\mathbm{0.42}$ & $\mathbm{0.40}$ & $\mathbm{0.52}$ \\
\cmidrule(lr){2-9}
 & \multirow{2}{*}{like} & \g $\mathcal{A}_{\mathrm{comp}}$ & \g $0.32$ & \g $\mathbm{0.41}$ & \g $\mathbm{0.39}$ & \g $\mathbm{0.55}$ & \g $\mathbm{0.55}$ & \g $\mathbm{0.57}$ \\
 &  & $\mathcal{A}_{\mathrm{exist}}$ & $0.23$ & $\mathbm{0.46}$ & $0.32$ & $\mathbm{0.42}$ & $\mathbm{0.46}$ & $\mathbm{0.49}$ \\
\cmidrule(lr){2-9}
 & \multirow{2}{*}{swap} & \g $\mathcal{A}_{\mathrm{comp}}$ & \g $0.15$ & \g $0.16$ & \g $0.18$ & \g $0.20$ & \g $0.20$ & \g $0.23$ \\
 &  & $\mathcal{A}_{\mathrm{exist}}$ & $0.09$ & $0.15$ & $0.12$ & $0.12$ & $0.14$ & $0.15$ \\
\cmidrule(lr){2-9}
 & \multirow{2}{*}{obf} & \g $\mathcal{A}_{\mathrm{comp}}$ & \g $0.19$ & \g $0.25$ & \g $0.21$ & \g $0.13$ & \g $0.12$ & \g $0.19$ \\
 &  & $\mathcal{A}_{\mathrm{exist}}$ & $0.19$ & $0.19$ & $\mathbm{0.26}$ & $\mathbm{0.30}$ & $0.18$ & $\mathbm{0.24}$ \\
\midrule
\multirow{8}{*}{\texttt{blood}} & \multirow{2}{*}{real} & \g $\mathcal{A}_{\mathrm{comp}}$ & \g $0.27$ & \g $0.21$ & \g $0.17$ & \g $0.24$ & \g $0.25$ & \g $0.27$ \\
 &  & $\mathcal{A}_{\mathrm{exist}}$ & $0.18$ & $0.21$ & $0.20$ & $0.28$ & $0.28$ & $0.28$ \\
\cmidrule(lr){2-9}
 & \multirow{2}{*}{like} & \g $\mathcal{A}_{\mathrm{comp}}$ & \g $0.23$ & \g $0.22$ & \g $0.20$ & \g $0.21$ & \g $0.21$ & \g $0.20$ \\
 &  & $\mathcal{A}_{\mathrm{exist}}$ & $0.18$ & $0.23$ & $0.23$ & $0.21$ & $0.23$ & $0.26$ \\
\cmidrule(lr){2-9}
 & \multirow{2}{*}{swap} & \g $\mathcal{A}_{\mathrm{comp}}$ & \g $0.20$ & \g $0.20^{*}$ & \g $0.28^{*}$ & \g $0.17$ & \g $0.23^{*}$ & \g $0.23^{*}$ \\
 &  & $\mathcal{A}_{\mathrm{exist}}$ & $0.15$ & $0.23^{*}$ & $0.26^{*}$ & $0.19^{*}$ & $0.16$ & $0.20^{*}$ \\
\cmidrule(lr){2-9}
 & \multirow{2}{*}{obf} & \g $\mathcal{A}_{\mathrm{comp}}$ & \g $0.23$ & \g $\mathbm{0.30}$ & \g $0.23$ & \g $0.25$ & \g $\mathbm{0.28}$ & \g $\mathbm{0.31}$ \\
 &  & $\mathcal{A}_{\mathrm{exist}}$ & $0.21$ & $0.23$ & $0.27$ & $0.22$ & $0.17$ & $0.19$ \\
\midrule
\multirow{8}{*}{\texttt{credit}} & \multirow{2}{*}{real} & \g $\mathcal{A}_{\mathrm{comp}}$ & \g $0.20$ & \g $\mathbm{0.52}$ & \g $\mathbm{0.35}$ & \g $\mathbm{0.69}$ & \g $\mathbm{0.81}^{*}$ & \g $\mathbm{0.39}$ \\
 &  & $\mathcal{A}_{\mathrm{exist}}$ & $0.21$ & $\mathbm{0.40}$ & $0.28$ & $\mathbm{0.46}$ & $\mathbm{0.37}$ & $0.14$ \\
\cmidrule(lr){2-9}
 & \multirow{2}{*}{like} & \g $\mathcal{A}_{\mathrm{comp}}$ & \g $0.21$ & \g $0.29$ & \g $0.30$ & \g $\mathbm{0.41}$ & \g $\mathbm{0.37}$ & \g $0.23$ \\
 &  & $\mathcal{A}_{\mathrm{exist}}$ & $0.21$ & $0.29$ & $0.29$ & $\mathbm{0.52}$ & $0.28$ & $0.20$ \\
\cmidrule(lr){2-9}
 & \multirow{2}{*}{swap} & \g $\mathcal{A}_{\mathrm{comp}}$ & \g $0.20^{*}$ & \g $0.23^{*}$ & \g $0.23^{*}$ & \g $0.17$ & \g $0.23^{*}$ & \g $0.32^{*}$ \\
 &  & $\mathcal{A}_{\mathrm{exist}}$ & $0.17$ & $0.17$ & $0.25^{*}$ & $0.24^{*}$ & $0.17$ & $0.21^{*}$ \\
\cmidrule(lr){2-9}
 & \multirow{2}{*}{obf} & \g $\mathcal{A}_{\mathrm{comp}}$ & \g $\mathbm{0.36}$ & \g $\mathbm{0.60}$ & \g $\mathbm{0.53}$ & \g $\mathbm{0.53}$ & \g $\mathbm{0.53}$ & \g $\mathbm{0.68}$ \\
 &  & $\mathcal{A}_{\mathrm{exist}}$ & $0.18$ & $\mathbm{0.32}$ & $\mathbm{0.35}$ & $\mathbm{0.29}$ & $0.23$ & $0.24$ \\
\midrule
\multirow{8}{*}{\texttt{diabetes}} & \multirow{2}{*}{real} & \g $\mathcal{A}_{\mathrm{comp}}$ & \g $0.21$ & \g $0.17$ & \g $0.17$ & \g $0.21$ & \g $0.23$ & \g $0.18$ \\
 &  & $\mathcal{A}_{\mathrm{exist}}$ & $0.20$ & $0.19$ & $0.16$ & $0.28$ & $0.24$ & $0.17$ \\
\cmidrule(lr){2-9}
 & \multirow{2}{*}{like} & \g $\mathcal{A}_{\mathrm{comp}}$ & \g $0.17$ & \g $0.19$ & \g $0.15$ & \g $0.23$ & \g $0.20$ & \g $0.18$ \\
 &  & $\mathcal{A}_{\mathrm{exist}}$ & $0.30$ & $0.15$ & $0.15$ & $0.21$ & $0.20$ & $0.18$ \\
\cmidrule(lr){2-9}
 & \multirow{2}{*}{swap} & \g $\mathcal{A}_{\mathrm{comp}}$ & \g $0.15$ & \g $0.18$ & \g $0.20$ & \g $0.15$ & \g $0.14$ & \g $0.14$ \\
 &  & $\mathcal{A}_{\mathrm{exist}}$ & $0.11$ & $0.21$ & $0.14$ & $0.12$ & $0.14$ & $0.17$ \\
\cmidrule(lr){2-9}
 & \multirow{2}{*}{obf} & \g $\mathcal{A}_{\mathrm{comp}}$ & \g $0.19$ & \g $\mathbm{0.41}$ & \g $\mathbm{0.36}$ & \g $0.29$ & \g $0.22$ & \g $\mathbm{0.42}$ \\
 &  & $\mathcal{A}_{\mathrm{exist}}$ & $0.23$ & $0.22$ & $0.24$ & $0.21$ & $0.20$ & $0.18$ \\
\midrule
\multirow{8}{*}{\texttt{gamma}} & \multirow{2}{*}{real} & \g $\mathcal{A}_{\mathrm{comp}}$ & \g $0.21$ & \g $0.13$ & \g $0.17$ & \g $0.20$ & \g $0.14$ & \g $0.20$ \\
 &  & $\mathcal{A}_{\mathrm{exist}}$ & $0.17$ & $0.17$ & $0.17$ & $0.14$ & $0.11$ & $0.18$ \\
\cmidrule(lr){2-9}
 & \multirow{2}{*}{like} & \g $\mathcal{A}_{\mathrm{comp}}$ & \g $0.20$ & \g $0.20$ & \g $0.20$ & \g $0.19$ & \g $0.17$ & \g $0.17$ \\
 &  & $\mathcal{A}_{\mathrm{exist}}$ & $0.17$ & $0.23$ & $0.19$ & $0.18$ & $0.20$ & $0.23$ \\
\cmidrule(lr){2-9}
 & \multirow{2}{*}{swap} & \g $\mathcal{A}_{\mathrm{comp}}$ & \g $0.20$ & \g $0.14$ & \g $0.23$ & \g $0.17$ & \g $0.17$ & \g $0.22$ \\
 &  & $\mathcal{A}_{\mathrm{exist}}$ & $0.14$ & $0.15$ & $0.18$ & $0.20$ & $0.15$ & $0.20$ \\
\cmidrule(lr){2-9}
 & \multirow{2}{*}{obf} & \g $\mathcal{A}_{\mathrm{comp}}$ & \g $0.21$ & \g $0.17$ & \g $0.21$ & \g $0.14$ & \g $0.18$ & \g $0.23^{*}$ \\
 &  & $\mathcal{A}_{\mathrm{exist}}$ & $0.15$ & $0.14$ & $0.15$ & $0.15$ & $0.14$ & $0.15$ \\
\midrule
\multirow{8}{*}{\texttt{iris}} & \multirow{2}{*}{real} & \g $\mathcal{A}_{\mathrm{comp}}$ & \g $0.15$ & \g $\mathbm{0.40}^{*}$ & \g $0.35^{*}$ & \g $\mathbm{0.47}^{*}$ & \g $\mathbm{0.47}^{*}$ & \g $\mathbm{0.53}^{*}$ \\
 &  & $\mathcal{A}_{\mathrm{exist}}$ & $0.21$ & $0.24$ & $0.28$ & $0.35$ & $0.36$ & $0.36$ \\
\cmidrule(lr){2-9}
 & \multirow{2}{*}{like} & \g $\mathcal{A}_{\mathrm{comp}}$ & \g $0.24$ & \g $0.33$ & \g $0.31$ & \g $0.19$ & \g $0.32$ & \g $0.25$ \\
 &  & $\mathcal{A}_{\mathrm{exist}}$ & $0.20$ & $0.31$ & $0.15$ & $0.20$ & $0.29$ & $0.20$ \\
\cmidrule(lr){2-9}
 & \multirow{2}{*}{swap} & \g $\mathcal{A}_{\mathrm{comp}}$ & \g $0.17$ & \g $0.16$ & \g $0.08$ & \g $0.15$ & \g $0.12$ & \g $0.12$ \\
 &  & $\mathcal{A}_{\mathrm{exist}}$ & $0.19$ & $0.15$ & $0.19$ & $0.25$ & $0.12$ & $0.20$ \\
\cmidrule(lr){2-9}
 & \multirow{2}{*}{obf} & \g $\mathcal{A}_{\mathrm{comp}}$ & \g $0.23$ & \g $0.24$ & \g $0.24$ & \g $0.19$ & \g $0.20$ & \g $0.17$ \\
 &  & $\mathcal{A}_{\mathrm{exist}}$ & $0.17$ & $0.19$ & $0.11$ & $0.28$ & $0.16$ & $0.21$ \\
\midrule
\multirow{8}{*}{\texttt{mushroom}} & \multirow{2}{*}{real} & \g $\mathcal{A}_{\mathrm{comp}}$ & \g $0.15$ & \g $0.23$ & \g $0.21$ & \g $0.25$ & \g $0.20$ & \g $0.15$ \\
 &  & $\mathcal{A}_{\mathrm{exist}}$ & $0.15$ & $0.16$ & $0.20$ & $0.17$ & $0.17$ & $0.14$ \\
\cmidrule(lr){2-9}
 & \multirow{2}{*}{like} & \g $\mathcal{A}_{\mathrm{comp}}$ & \g $0.20$ & \g $0.21$ & \g $0.17$ & \g $0.12$ & \g $0.17$ & \g $0.12$ \\
 &  & $\mathcal{A}_{\mathrm{exist}}$ & $0.18$ & $0.14$ & $0.19$ & $0.14$ & $0.11$ & $0.18$ \\
\cmidrule(lr){2-9}
 & \multirow{2}{*}{swap} & \g $\mathcal{A}_{\mathrm{comp}}$ & \g $0.14$ & \g $0.12$ & \g $0.10$ & \g $0.09$ & \g $0.14$ & \g $0.12$ \\
 &  & $\mathcal{A}_{\mathrm{exist}}$ & $0.13$ & $0.10$ & $0.12$ & $0.07$ & $0.10$ & $0.09$ \\
\cmidrule(lr){2-9}
 & \multirow{2}{*}{obf} & \g $\mathcal{A}_{\mathrm{comp}}$ & \g $0.20$ & \g $0.14$ & \g $0.14$ & \g $0.12$ & \g $0.17$ & \g $0.12$ \\
 &  & $\mathcal{A}_{\mathrm{exist}}$ & $0.17$ & $0.18$ & $0.17$ & $0.14$ & $0.15$ & $0.15$ \\
\midrule
\multirow{8}{*}{\texttt{synthetic}} & \multirow{2}{*}{real} & \g $\mathcal{A}_{\mathrm{comp}}$ & \g $0.15$ & \g $0.17$ & \g $0.15$ & \g $0.14$ & \g $0.10$ & \g $0.18$ \\
 &  & $\mathcal{A}_{\mathrm{exist}}$ & $0.19$ & $0.16$ & $0.14$ & $0.15$ & $0.14$ & $0.19$ \\
\cmidrule(lr){2-9}
 & \multirow{2}{*}{like} & \g $\mathcal{A}_{\mathrm{comp}}$ & \g $0.23$ & \g $0.24$ & \g $0.20$ & \g $0.18$ & \g $0.20$ & \g $0.20$ \\
 &  & $\mathcal{A}_{\mathrm{exist}}$ & $0.19$ & $0.21$ & $0.21$ & $0.17$ & $0.14$ & $0.18$ \\
\cmidrule(lr){2-9}
 & \multirow{2}{*}{swap} & \g $\mathcal{A}_{\mathrm{comp}}$ & \g $0.17$ & \g $0.17$ & \g $0.15$ & \g $0.09$ & \g $0.10$ & \g $0.17$ \\
 &  & $\mathcal{A}_{\mathrm{exist}}$ & $0.18$ & $0.14$ & $0.18$ & $0.20$ & $0.17$ & $0.20$ \\
\cmidrule(lr){2-9}
 & \multirow{2}{*}{obf} & \g $\mathcal{A}_{\mathrm{comp}}$ & \g $0.14$ & \g $0.14$ & \g $0.09$ & \g $0.10$ & \g $0.10$ & \g $0.13$ \\
 &  & $\mathcal{A}_{\mathrm{exist}}$ & $0.17$ & $0.16$ & $0.23$ & $0.18$ & $0.20$ & $0.20$ \\
\midrule
\multirow{8}{*}{\texttt{titanic}} & \multirow{2}{*}{real} & \g $\mathcal{A}_{\mathrm{comp}}$ & \g $\mathbm{0.32}$ & \g $\mathbm{0.47}^{*}$ & \g $\mathbm{0.43}$ & \g $\mathbm{0.54}^{*}$ & \g $\mathbm{0.50}^{*}$ & \g $\mathbm{0.57}^{*}$ \\
 &  & $\mathcal{A}_{\mathrm{exist}}$ & $0.17$ & $\mathbm{0.34}$ & $0.20$ & $\mathbm{0.39}^{*}$ & $\mathbm{0.31}$ & $\mathbm{0.42}^{*}$ \\
\cmidrule(lr){2-9}
 & \multirow{2}{*}{like} & \g $\mathcal{A}_{\mathrm{comp}}$ & \g $\mathbm{0.33}$ & \g $\mathbm{0.40}$ & \g $\mathbm{0.32}$ & \g $\mathbm{0.47}^{*}$ & \g $\mathbm{0.40}$ & \g $\mathbm{0.54}^{*}$ \\
 &  & $\mathcal{A}_{\mathrm{exist}}$ & $0.19$ & $\mathbm{0.32}$ & $0.18$ & $\mathbm{0.36}$ & $0.24$ & $\mathbm{0.41}$ \\
\cmidrule(lr){2-9}
 & \multirow{2}{*}{swap} & \g $\mathcal{A}_{\mathrm{comp}}$ & \g $\mathbm{0.33}^{*}$ & \g $0.19$ & \g $0.26^{*}$ & \g $\mathbm{0.29}^{*}$ & \g $0.18$ & \g $0.12$ \\
 &  & $\mathcal{A}_{\mathrm{exist}}$ & $0.19$ & $0.23$ & $0.10$ & $0.25$ & $0.13$ & $0.17$ \\
\cmidrule(lr){2-9}
 & \multirow{2}{*}{obf} & \g $\mathcal{A}_{\mathrm{comp}}$ & \g $0.20$ & \g $0.21$ & \g $0.27$ & \g $0.30$ & \g $0.25$ & \g $0.28$ \\
 &  & $\mathcal{A}_{\mathrm{exist}}$ & $0.26$ & $0.23$ & $0.19$ & $0.29$ & $0.27$ & $0.29$ \\
\end{xltabular}

\keepXColumns
\begin{xltabular}{0.9\textwidth}{l l l c c c c c}
\caption{\textbf{Summary of contamination results across datasets and models (temperature = 0.6, coverage = 20\%).} $\mathcal{A}_{\mathrm{comp}}$ denotes completion accuracy, and $\mathcal{A}_{\mathrm{exist}}$ denotes existence accuracy. Bold values indicate performances that are statistically significant above the random baseline and have higher accuracy than the random baseline ($\alpha = 0.01$); $^{*}$ indicates significance against the deterministic baseline, the stochastic baseline, or both ($\alpha = 0.01$).}\label{tab:contamination_summary_t0.6_cov20} \\
\toprule
\multirow{2}{*}{\centering\textbf{Dataset}} & \multirow{2}{*}{\centering\textbf{Variant}} & \multirow{2}{*}{\centering\textbf{Metric}} & \multicolumn{5}{c}{\textbf{Model}} \\
\cmidrule(lr){4-8}
 &  &  & \textit{Mistral-7B} & \textit{Qwen-7B} & \textit{Llama-8B} & \textit{Qwen-14B} & \textit{Qwen-32B} \\
\midrule
\endfirsthead
\caption[]{Summary of contamination results across datasets and models (continued)} \\
\toprule
\multirow{2}{*}{\centering\textbf{Dataset}} & \multirow{2}{*}{\centering\textbf{Variant}} & \multirow{2}{*}{\centering\textbf{Metric}} & \multicolumn{5}{c}{\textbf{Model}} \\
\cmidrule(lr){4-8}
 &  &  & \textit{Mistral-7B} & \textit{Qwen-7B} & \textit{Llama-8B} & \textit{Qwen-14B} & \textit{Qwen-32B} \\
\midrule
\endhead
\midrule
\multicolumn{8}{r}{\textit{Continued on next page}} \\
\midrule
\endfoot
\bottomrule
\endlastfoot
\multirow{8}{*}{\texttt{adult}} & \multirow{2}{*}{real} & \g $\mathcal{A}_{\mathrm{comp}}$ & \g $\mathbm{0.35}$ & \g $\mathbm{0.49}$ & \g $\mathbm{0.54}$ & \g $\mathbm{0.65}$ & \g $\mathbm{0.69}$ \\
 &  & $\mathcal{A}_{\mathrm{exist}}$ & $0.31$ & $0.32$ & $0.23$ & $\mathbm{0.44}$ & $\mathbm{0.43}$ \\
\cmidrule(lr){2-8}
 & \multirow{2}{*}{like} & \g $\mathcal{A}_{\mathrm{comp}}$ & \g $0.33$ & \g $\mathbm{0.46}$ & \g $\mathbm{0.37}$ & \g $\mathbm{0.49}$ & \g $\mathbm{0.56}$ \\
 &  & $\mathcal{A}_{\mathrm{exist}}$ & $0.30$ & $\mathbm{0.45}$ & $0.28$ & $\mathbm{0.46}$ & $\mathbm{0.46}$ \\
\cmidrule(lr){2-8}
 & \multirow{2}{*}{swap} & \g $\mathcal{A}_{\mathrm{comp}}$ & \g $0.17$ & \g $0.17$ & \g $0.16$ & \g $0.15$ & \g $0.20$ \\
 &  & $\mathcal{A}_{\mathrm{exist}}$ & $0.18$ & $0.17$ & $0.18$ & $0.10$ & $0.10$ \\
\cmidrule(lr){2-8}
 & \multirow{2}{*}{obf} & \g $\mathcal{A}_{\mathrm{comp}}$ & \g $0.20$ & \g $0.24$ & \g $0.24$ & \g $0.14$ & \g $0.09$ \\
 &  & $\mathcal{A}_{\mathrm{exist}}$ & $\mathbm{0.28}$ & $0.22$ & $\mathbm{0.29}$ & $\mathbm{0.24}$ & $0.17$ \\
\midrule
\multirow{8}{*}{\texttt{blood}} & \multirow{2}{*}{real} & \g $\mathcal{A}_{\mathrm{comp}}$ & \g $0.21$ & \g $0.18$ & \g $0.19$ & \g $0.21$ & \g $0.28$ \\
 &  & $\mathcal{A}_{\mathrm{exist}}$ & $0.21$ & $0.28$ & $0.16$ & $0.28$ & $0.23$ \\
\cmidrule(lr){2-8}
 & \multirow{2}{*}{like} & \g $\mathcal{A}_{\mathrm{comp}}$ & \g $0.26$ & \g $0.22$ & \g $0.24$ & \g $0.20$ & \g $0.23$ \\
 &  & $\mathcal{A}_{\mathrm{exist}}$ & $0.22$ & $0.20$ & $0.25$ & $0.27$ & $0.24$ \\
\cmidrule(lr){2-8}
 & \multirow{2}{*}{swap} & \g $\mathcal{A}_{\mathrm{comp}}$ & \g $0.17$ & \g $0.20^{*}$ & \g $0.17$ & \g $0.23^{*}$ & \g $0.21^{*}$ \\
 &  & $\mathcal{A}_{\mathrm{exist}}$ & $0.14$ & $0.23^{*}$ & $0.21^{*}$ & $0.17^{*}$ & $0.20^{*}$ \\
\cmidrule(lr){2-8}
 & \multirow{2}{*}{obf} & \g $\mathcal{A}_{\mathrm{comp}}$ & \g $0.20$ & \g $0.25$ & \g $0.18$ & \g $\mathbm{0.28}$ & \g $0.23$ \\
 &  & $\mathcal{A}_{\mathrm{exist}}$ & $0.17$ & $0.19$ & $0.18$ & $0.19$ & $0.14$ \\
\midrule
\multirow{8}{*}{\texttt{credit}} & \multirow{2}{*}{real} & \g $\mathcal{A}_{\mathrm{comp}}$ & \g $0.24$ & \g $\mathbm{0.53}$ & \g $\mathbm{0.38}$ & \g $\mathbm{0.66}$ & \g $\mathbm{0.76}^{*}$ \\
 &  & $\mathcal{A}_{\mathrm{exist}}$ & $0.22$ & $\mathbm{0.39}$ & $0.23$ & $\mathbm{0.46}$ & $\mathbm{0.39}$ \\
\cmidrule(lr){2-8}
 & \multirow{2}{*}{like} & \g $\mathcal{A}_{\mathrm{comp}}$ & \g $0.23$ & \g $0.32$ & \g $0.23$ & \g $\mathbm{0.41}$ & \g $\mathbm{0.37}$ \\
 &  & $\mathcal{A}_{\mathrm{exist}}$ & $0.28$ & $0.28$ & $0.27$ & $\mathbm{0.45}$ & $0.27$ \\
\cmidrule(lr){2-8}
 & \multirow{2}{*}{swap} & \g $\mathcal{A}_{\mathrm{comp}}$ & \g $0.21^{*}$ & \g $0.20$ & \g $0.18$ & \g $0.15$ & \g $0.19$ \\
 &  & $\mathcal{A}_{\mathrm{exist}}$ & $0.18$ & $0.20$ & $0.19$ & $0.23^{*}$ & $0.19$ \\
\cmidrule(lr){2-8}
 & \multirow{2}{*}{obf} & \g $\mathcal{A}_{\mathrm{comp}}$ & \g $\mathbm{0.35}$ & \g $\mathbm{0.66}$ & \g $\mathbm{0.50}$ & \g $\mathbm{0.56}$ & \g $\mathbm{0.57}$ \\
 &  & $\mathcal{A}_{\mathrm{exist}}$ & $\mathbm{0.30}$ & $\mathbm{0.34}$ & $\mathbm{0.31}$ & $\mathbm{0.30}$ & $0.26$ \\
\midrule
\multirow{8}{*}{\texttt{diabetes}} & \multirow{2}{*}{real} & \g $\mathcal{A}_{\mathrm{comp}}$ & \g $0.19$ & \g $0.21$ & \g $0.26$ & \g $0.19$ & \g $0.23$ \\
 &  & $\mathcal{A}_{\mathrm{exist}}$ & $0.26$ & $0.18$ & $0.20$ & $0.24$ & $0.27$ \\
\cmidrule(lr){2-8}
 & \multirow{2}{*}{like} & \g $\mathcal{A}_{\mathrm{comp}}$ & \g $0.20$ & \g $0.18$ & \g $0.23$ & \g $0.24$ & \g $0.17$ \\
 &  & $\mathcal{A}_{\mathrm{exist}}$ & $0.31$ & $0.14$ & $0.17$ & $0.23$ & $0.21$ \\
\cmidrule(lr){2-8}
 & \multirow{2}{*}{swap} & \g $\mathcal{A}_{\mathrm{comp}}$ & \g $0.18$ & \g $0.23^{*}$ & \g $0.15$ & \g $0.17$ & \g $0.12$ \\
 &  & $\mathcal{A}_{\mathrm{exist}}$ & $0.12$ & $0.16$ & $0.17$ & $0.15$ & $0.14$ \\
\cmidrule(lr){2-8}
 & \multirow{2}{*}{obf} & \g $\mathcal{A}_{\mathrm{comp}}$ & \g $0.22$ & \g $\mathbm{0.41}$ & \g $\mathbm{0.39}$ & \g $0.27$ & \g $0.21$ \\
 &  & $\mathcal{A}_{\mathrm{exist}}$ & $0.23$ & $0.24$ & $0.23$ & $0.22$ & $0.20$ \\
\midrule
\multirow{8}{*}{\texttt{gamma}} & \multirow{2}{*}{real} & \g $\mathcal{A}_{\mathrm{comp}}$ & \g $0.18$ & \g $0.19$ & \g $0.24^{*}$ & \g $0.17$ & \g $0.17$ \\
 &  & $\mathcal{A}_{\mathrm{exist}}$ & $0.18$ & $0.20^{*}$ & $0.19$ & $0.11$ & $0.14$ \\
\cmidrule(lr){2-8}
 & \multirow{2}{*}{like} & \g $\mathcal{A}_{\mathrm{comp}}$ & \g $0.19$ & \g $0.16$ & \g $0.20$ & \g $0.19$ & \g $0.20$ \\
 &  & $\mathcal{A}_{\mathrm{exist}}$ & $0.18$ & $0.20$ & $0.18$ & $0.17$ & $0.19$ \\
\cmidrule(lr){2-8}
 & \multirow{2}{*}{swap} & \g $\mathcal{A}_{\mathrm{comp}}$ & \g $0.18$ & \g $0.14$ & \g $0.21$ & \g $0.15$ & \g $0.17$ \\
 &  & $\mathcal{A}_{\mathrm{exist}}$ & $0.17$ & $0.19$ & $0.20$ & $0.23$ & $0.17$ \\
\cmidrule(lr){2-8}
 & \multirow{2}{*}{obf} & \g $\mathcal{A}_{\mathrm{comp}}$ & \g $0.20$ & \g $0.20$ & \g $0.18$ & \g $0.15$ & \g $0.21$ \\
 &  & $\mathcal{A}_{\mathrm{exist}}$ & $0.19$ & $0.14$ & $0.18$ & $0.16$ & $0.19$ \\
\midrule
\multirow{8}{*}{\texttt{iris}} & \multirow{2}{*}{real} & \g $\mathcal{A}_{\mathrm{comp}}$ & \g $0.28$ & \g $0.32$ & \g $0.33^{*}$ & \g $\mathbm{0.43}^{*}$ & \g $\mathbm{0.56}^{*}$ \\
 &  & $\mathcal{A}_{\mathrm{exist}}$ & $0.20$ & $0.21$ & $0.23$ & $0.33$ & $\mathbm{0.47}^{*}$ \\
\cmidrule(lr){2-8}
 & \multirow{2}{*}{like} & \g $\mathcal{A}_{\mathrm{comp}}$ & \g $0.20$ & \g $0.23$ & \g $0.24$ & \g $0.24$ & \g $0.28$ \\
 &  & $\mathcal{A}_{\mathrm{exist}}$ & $0.28$ & $0.19$ & $0.20$ & $0.21$ & $0.20$ \\
\cmidrule(lr){2-8}
 & \multirow{2}{*}{swap} & \g $\mathcal{A}_{\mathrm{comp}}$ & \g $0.15$ & \g $0.20$ & \g $0.11$ & \g $0.15$ & \g $0.11$ \\
 &  & $\mathcal{A}_{\mathrm{exist}}$ & $0.23$ & $0.24$ & $0.12$ & $0.28$ & $0.23$ \\
\cmidrule(lr){2-8}
 & \multirow{2}{*}{obf} & \g $\mathcal{A}_{\mathrm{comp}}$ & \g $0.17$ & \g $0.20$ & \g $0.27$ & \g $0.20$ & \g $0.12$ \\
 &  & $\mathcal{A}_{\mathrm{exist}}$ & $0.27$ & $0.23$ & $0.17$ & $0.29$ & $0.29$ \\
\midrule
\multirow{8}{*}{\texttt{mushroom}} & \multirow{2}{*}{real} & \g $\mathcal{A}_{\mathrm{comp}}$ & \g $0.18$ & \g $0.18$ & \g $0.17$ & \g $0.17$ & \g $0.21$ \\
 &  & $\mathcal{A}_{\mathrm{exist}}$ & $0.20$ & $0.14$ & $0.17$ & $0.17$ & $0.12$ \\
\cmidrule(lr){2-8}
 & \multirow{2}{*}{like} & \g $\mathcal{A}_{\mathrm{comp}}$ & \g $\mathbm{0.26}$ & \g $0.23$ & \g $0.20$ & \g $0.12$ & \g $0.19$ \\
 &  & $\mathcal{A}_{\mathrm{exist}}$ & $0.20$ & $0.16$ & $0.17$ & $0.13$ & $0.10$ \\
\cmidrule(lr){2-8}
 & \multirow{2}{*}{swap} & \g $\mathcal{A}_{\mathrm{comp}}$ & \g $0.14$ & \g $0.14$ & \g $0.12$ & \g $0.04$ & \g $0.10$ \\
 &  & $\mathcal{A}_{\mathrm{exist}}$ & $0.14$ & $0.15$ & $0.10$ & $0.08$ & $0.12$ \\
\cmidrule(lr){2-8}
 & \multirow{2}{*}{obf} & \g $\mathcal{A}_{\mathrm{comp}}$ & \g $0.18$ & \g $0.10$ & \g $0.15$ & \g $0.13$ & \g $0.12$ \\
 &  & $\mathcal{A}_{\mathrm{exist}}$ & $0.15$ & $0.18$ & $0.17$ & $0.20$ & $0.15$ \\
\midrule
\multirow{8}{*}{\texttt{synthetic}} & \multirow{2}{*}{real} & \g $\mathcal{A}_{\mathrm{comp}}$ & \g $0.15$ & \g $0.15$ & \g $0.17$ & \g $0.11$ & \g $0.14$ \\
 &  & $\mathcal{A}_{\mathrm{exist}}$ & $0.17$ & $0.17$ & $0.17$ & $0.15$ & $0.15$ \\
\cmidrule(lr){2-8}
 & \multirow{2}{*}{like} & \g $\mathcal{A}_{\mathrm{comp}}$ & \g $0.16$ & \g $0.21$ & \g $0.23$ & \g $0.20$ & \g $0.20$ \\
 &  & $\mathcal{A}_{\mathrm{exist}}$ & $0.24$ & $0.17$ & $0.21$ & $0.20$ & $0.14$ \\
\cmidrule(lr){2-8}
 & \multirow{2}{*}{swap} & \g $\mathcal{A}_{\mathrm{comp}}$ & \g $0.17$ & \g $0.13$ & \g $0.20$ & \g $0.10$ & \g $0.14$ \\
 &  & $\mathcal{A}_{\mathrm{exist}}$ & $0.19$ & $0.17$ & $0.17$ & $0.20$ & $0.17$ \\
\cmidrule(lr){2-8}
 & \multirow{2}{*}{obf} & \g $\mathcal{A}_{\mathrm{comp}}$ & \g $0.10$ & \g $0.16$ & \g $0.18$ & \g $0.12$ & \g $0.14$ \\
 &  & $\mathcal{A}_{\mathrm{exist}}$ & $0.20$ & $0.17$ & $0.25$ & $0.17$ & $0.15$ \\
\midrule
\multirow{8}{*}{\texttt{titanic}} & \multirow{2}{*}{real} & \g $\mathcal{A}_{\mathrm{comp}}$ & \g $\mathbm{0.29}$ & \g $\mathbm{0.47}^{*}$ & \g $\mathbm{0.38}$ & \g $\mathbm{0.49}^{*}$ & \g $\mathbm{0.43}^{*}$ \\
 &  & $\mathcal{A}_{\mathrm{exist}}$ & $0.20$ & $\mathbm{0.30}$ & $0.21$ & $\mathbm{0.40}^{*}$ & $0.26$ \\
\cmidrule(lr){2-8}
 & \multirow{2}{*}{like} & \g $\mathcal{A}_{\mathrm{comp}}$ & \g $0.28$ & \g $\mathbm{0.40}$ & \g $\mathbm{0.32}$ & \g $\mathbm{0.41}$ & \g $\mathbm{0.43}$ \\
 &  & $\mathcal{A}_{\mathrm{exist}}$ & $0.25$ & $0.28$ & $0.20$ & $\mathbm{0.35}$ & $0.28$ \\
\cmidrule(lr){2-8}
 & \multirow{2}{*}{swap} & \g $\mathcal{A}_{\mathrm{comp}}$ & \g $0.23$ & \g $0.20$ & \g $0.20$ & \g $0.24$ & \g $0.18$ \\
 &  & $\mathcal{A}_{\mathrm{exist}}$ & $0.22$ & $0.28$ & $0.14$ & $0.28$ & $0.19$ \\
\cmidrule(lr){2-8}
 & \multirow{2}{*}{obf} & \g $\mathcal{A}_{\mathrm{comp}}$ & \g $0.24$ & \g $0.21$ & \g $0.26$ & \g $0.24$ & \g $0.26$ \\
 &  & $\mathcal{A}_{\mathrm{exist}}$ & $0.28$ & $0.23$ & $0.17$ & $0.28$ & $0.22$ \\
\end{xltabular}

\keepXColumns
\begin{xltabular}{0.9\textwidth}{l l l c c c c c}
\caption{\textbf{Summary of contamination results across datasets and models (temperature = 0.0, coverage = 50\%)}. $\mathcal{A}_{\mathrm{comp}}$ denotes completion accuracy, and $\mathcal{A}_{\mathrm{exist}}$ denotes existence accuracy. Bold values indicate performances that are statistically significant above the random baseline and have higher accuracy than the random baseline ($\alpha = 0.01$); $^{*}$ indicates significance against the deterministic baseline, the stochastic baseline, or both ($\alpha = 0.01$).}\label{tab:contamination_summary_t0_cov50} \\
\toprule
\multirow{2}{*}{\centering\textbf{Dataset}} & \multirow{2}{*}{\centering\textbf{Variant}} & \multirow{2}{*}{\centering\textbf{Metric}} & \multicolumn{5}{c}{\textbf{Model}} \\
\cmidrule(lr){4-8}
 &  &  & \textit{Mistral-7B} & \textit{Qwen-7B} & \textit{Llama-8B} & \textit{Qwen-14B} & \textit{Qwen-32B} \\
\midrule
\endfirsthead
\caption[]{Summary of contamination results across datasets and models (continued)} \\
\toprule
\multirow{2}{*}{\centering\textbf{Dataset}} & \multirow{2}{*}{\centering\textbf{Variant}} & \multirow{2}{*}{\centering\textbf{Metric}} & \multicolumn{5}{c}{\textbf{Model}} \\
\cmidrule(lr){4-8}
 &  &  & \textit{Mistral-7B} & \textit{Qwen-7B} & \textit{Llama-8B} & \textit{Qwen-14B} & \textit{Qwen-32B} \\
\midrule
\endhead
\midrule
\multicolumn{8}{r}{\textit{Continued on next page}} \\
\midrule
\endfoot
\bottomrule
\endlastfoot
\multirow{8}{*}{\texttt{adult}} & \multirow{2}{*}{real} & \g $\mathcal{A}_{\mathrm{comp}}$ & \g $0.28$ & \g $\mathbm{0.55}$ & \g $\mathbm{0.47}$ & \g $\mathbm{0.76}$ & \g $\mathbm{0.79}$ \\
 &  & $\mathcal{A}_{\mathrm{exist}}$ & $0.23$ & $\mathbm{0.36}$ & $\mathbm{0.39}$ & $\mathbm{0.45}$ & $\mathbm{0.49}$ \\
\cmidrule(lr){2-8}
 & \multirow{2}{*}{like} & \g $\mathcal{A}_{\mathrm{comp}}$ & \g $\mathbm{0.36}$ & \g $\mathbm{0.51}$ & \g $\mathbm{0.47}$ & \g $\mathbm{0.65}$ & \g $\mathbm{0.65}$ \\
 &  & $\mathcal{A}_{\mathrm{exist}}$ & $0.24$ & $\mathbm{0.54}$ & $\mathbm{0.34}$ & $\mathbm{0.53}$ & $\mathbm{0.52}$ \\
\cmidrule(lr){2-8}
 & \multirow{2}{*}{swap} & \g $\mathcal{A}_{\mathrm{comp}}$ & \g $0.15$ & \g $0.10$ & \g $0.09$ & \g $0.19$ & \g $0.12$ \\
 &  & $\mathcal{A}_{\mathrm{exist}}$ & $0.14$ & $0.16$ & $0.23$ & $0.09$ & $0.15$ \\
\cmidrule(lr){2-8}
 & \multirow{2}{*}{obf} & \g $\mathcal{A}_{\mathrm{comp}}$ & \g $0.21$ & \g $0.21$ & \g $0.21$ & \g $0.07$ & \g $0.07$ \\
 &  & $\mathcal{A}_{\mathrm{exist}}$ & $0.17$ & $0.17$ & $0.21$ & $0.21$ & $0.23$ \\
\midrule
\multirow{8}{*}{\texttt{blood}} & \multirow{2}{*}{real} & \g $\mathcal{A}_{\mathrm{comp}}$ & \g $0.17$ & \g $0.24$ & \g $0.20$ & \g $0.28$ & \g $\mathbm{0.38}$ \\
 &  & $\mathcal{A}_{\mathrm{exist}}$ & $0.20$ & $0.22$ & $0.20$ & $0.35$ & $0.26$ \\
\cmidrule(lr){2-8}
 & \multirow{2}{*}{like} & \g $\mathcal{A}_{\mathrm{comp}}$ & \g $0.23$ & \g $0.25$ & \g $0.26$ & \g $0.27$ & \g $0.28$ \\
 &  & $\mathcal{A}_{\mathrm{exist}}$ & $0.23$ & $0.20$ & $0.17$ & $0.34$ & $0.24$ \\
\cmidrule(lr){2-8}
 & \multirow{2}{*}{swap} & \g $\mathcal{A}_{\mathrm{comp}}$ & \g $0.19^{*}$ & \g $0.20^{*}$ & \g $0.23^{*}$ & \g $0.14$ & \g $0.17$ \\
 &  & $\mathcal{A}_{\mathrm{exist}}$ & $0.15^{*}$ & $0.18^{*}$ & $0.22^{*}$ & $0.09$ & $0.11$ \\
\cmidrule(lr){2-8}
 & \multirow{2}{*}{obf} & \g $\mathcal{A}_{\mathrm{comp}}$ & \g $0.25$ & \g $0.24$ & \g $0.23$ & \g $\mathbm{0.28}$ & \g $\mathbm{0.33}$ \\
 &  & $\mathcal{A}_{\mathrm{exist}}$ & $0.26$ & $0.19$ & $0.26$ & $0.28$ & $0.23$ \\
\midrule
\multirow{8}{*}{\texttt{credit}} & \multirow{2}{*}{real} & \g $\mathcal{A}_{\mathrm{comp}}$ & \g $0.19$ & \g $0.20$ & \g $0.14$ & \g $0.20$ & \g $\mathbm{0.28}$ \\
 &  & $\mathcal{A}_{\mathrm{exist}}$ & $0.14$ & $0.21$ & $0.21$ & $\mathbm{0.35}$ & $\mathbm{0.34}$ \\
\cmidrule(lr){2-8}
 & \multirow{2}{*}{like} & \g $\mathcal{A}_{\mathrm{comp}}$ & \g $0.23$ & \g $0.19$ & \g $0.17$ & \g $\mathbm{0.41}$ & \g $0.28$ \\
 &  & $\mathcal{A}_{\mathrm{exist}}$ & $0.22$ & $0.32$ & $0.33$ & $\mathbm{0.56}$ & $\mathbm{0.51}$ \\
\cmidrule(lr){2-8}
 & \multirow{2}{*}{swap} & \g $\mathcal{A}_{\mathrm{comp}}$ & \g $0.24^{*}$ & \g $0.20^{*}$ & \g $0.28^{*}$ & \g $0.23^{*}$ & \g $0.24^{*}$ \\
 &  & $\mathcal{A}_{\mathrm{exist}}$ & $0.21^{*}$ & $0.27^{*}$ & $0.23^{*}$ & $0.22^{*}$ & $0.18^{*}$ \\
\cmidrule(lr){2-8}
 & \multirow{2}{*}{obf} & \g $\mathcal{A}_{\mathrm{comp}}$ & \g $0.21$ & \g $\mathbm{0.46}$ & \g $\mathbm{0.49}$ & \g $0.26$ & \g $0.30$ \\
 &  & $\mathcal{A}_{\mathrm{exist}}$ & $0.16$ & $0.27$ & $0.26$ & $0.24$ & $0.18$ \\
\midrule
\multirow{8}{*}{\texttt{diabetes}} & \multirow{2}{*}{real} & \g $\mathcal{A}_{\mathrm{comp}}$ & \g $0.23$ & \g $0.17$ & \g $0.23$ & \g $0.20$ & \g $0.28$ \\
 &  & $\mathcal{A}_{\mathrm{exist}}$ & $0.21$ & $0.17$ & $0.15$ & $0.24$ & $0.24$ \\
\cmidrule(lr){2-8}
 & \multirow{2}{*}{like} & \g $\mathcal{A}_{\mathrm{comp}}$ & \g $0.21$ & \g $0.18$ & \g $0.18$ & \g $0.21$ & \g $0.24$ \\
 &  & $\mathcal{A}_{\mathrm{exist}}$ & $0.26$ & $0.23$ & $0.11$ & $0.30$ & $0.16$ \\
\cmidrule(lr){2-8}
 & \multirow{2}{*}{swap} & \g $\mathcal{A}_{\mathrm{comp}}$ & \g $0.17$ & \g $0.23^{*}$ & \g $0.23^{*}$ & \g $0.17$ & \g $0.12$ \\
 &  & $\mathcal{A}_{\mathrm{exist}}$ & $0.16$ & $0.15$ & $0.18$ & $0.16$ & $0.12$ \\
\cmidrule(lr){2-8}
 & \multirow{2}{*}{obf} & \g $\mathcal{A}_{\mathrm{comp}}$ & \g $0.23$ & \g $\mathbm{0.36}$ & \g $\mathbm{0.33}$ & \g $\mathbm{0.30}$ & \g $0.25$ \\
 &  & $\mathcal{A}_{\mathrm{exist}}$ & $\mathbm{0.31}$ & $0.21$ & $0.27$ & $0.23$ & $0.17$ \\
\midrule
\multirow{8}{*}{\texttt{gamma}} & \multirow{2}{*}{real} & \g $\mathcal{A}_{\mathrm{comp}}$ & \g $0.17$ & \g $0.17^{*}$ & \g $0.14$ & \g $0.18^{*}$ & \g $0.19^{*}$ \\
 &  & $\mathcal{A}_{\mathrm{exist}}$ & $0.15$ & $0.14$ & $0.18$ & $0.14$ & $0.13$ \\
\cmidrule(lr){2-8}
 & \multirow{2}{*}{like} & \g $\mathcal{A}_{\mathrm{comp}}$ & \g $\mathbm{0.30}$ & \g $0.20$ & \g $0.19$ & \g $0.21$ & \g $\mathbm{0.27}$ \\
 &  & $\mathcal{A}_{\mathrm{exist}}$ & $0.23$ & $0.18$ & $0.21$ & $0.18$ & $0.14$ \\
\cmidrule(lr){2-8}
 & \multirow{2}{*}{swap} & \g $\mathcal{A}_{\mathrm{comp}}$ & \g $0.15$ & \g $0.14$ & \g $0.23$ & \g $0.20$ & \g $0.10$ \\
 &  & $\mathcal{A}_{\mathrm{exist}}$ & $0.15$ & $0.19$ & $0.21$ & $0.18$ & $0.16$ \\
\cmidrule(lr){2-8}
 & \multirow{2}{*}{obf} & \g $\mathcal{A}_{\mathrm{comp}}$ & \g $0.21^{*}$ & \g $0.10$ & \g $0.20$ & \g $0.17$ & \g $0.20$ \\
 &  & $\mathcal{A}_{\mathrm{exist}}$ & $0.17$ & $0.18^{*}$ & $0.18^{*}$ & $0.17^{*}$ & $0.15$ \\
\midrule
\multirow{8}{*}{\texttt{iris}} & \multirow{2}{*}{real} & \g $\mathcal{A}_{\mathrm{comp}}$ & \g $\mathbm{0.37}$ & \g $\mathbm{0.39}$ & \g $\mathbm{0.37}$ & \g $\mathbm{0.61}^{*}$ & \g $\mathbm{0.69}^{*}$ \\
 &  & $\mathcal{A}_{\mathrm{exist}}$ & $0.27$ & $0.36$ & $0.29$ & $\mathbm{0.48}^{*}$ & $\mathbm{0.51}^{*}$ \\
\cmidrule(lr){2-8}
 & \multirow{2}{*}{like} & \g $\mathcal{A}_{\mathrm{comp}}$ & \g $0.19$ & \g $0.24$ & \g $0.20$ & \g $0.16$ & \g $0.20$ \\
 &  & $\mathcal{A}_{\mathrm{exist}}$ & $0.09$ & $0.15$ & $0.25$ & $0.15$ & $0.17$ \\
\cmidrule(lr){2-8}
 & \multirow{2}{*}{swap} & \g $\mathcal{A}_{\mathrm{comp}}$ & \g $0.13$ & \g $0.21$ & \g $0.20$ & \g $0.21$ & \g $0.24$ \\
 &  & $\mathcal{A}_{\mathrm{exist}}$ & $0.16$ & $0.25$ & $0.12$ & $0.16$ & $0.23$ \\
\cmidrule(lr){2-8}
 & \multirow{2}{*}{obf} & \g $\mathcal{A}_{\mathrm{comp}}$ & \g $0.19$ & \g $0.13$ & \g $0.19$ & \g $0.16$ & \g $0.28$ \\
 &  & $\mathcal{A}_{\mathrm{exist}}$ & $0.16$ & $0.20$ & $0.15$ & $0.16$ & $0.17$ \\
\midrule
\multirow{8}{*}{\texttt{mushroom}} & \multirow{2}{*}{real} & \g $\mathcal{A}_{\mathrm{comp}}$ & \g $0.08$ & \g $0.18$ & \g $0.13$ & \g $0.12$ & \g $0.12$ \\
 &  & $\mathcal{A}_{\mathrm{exist}}$ & $0.12$ & $\mathbm{0.29}$ & $0.14$ & $0.20$ & $0.14$ \\
\cmidrule(lr){2-8}
 & \multirow{2}{*}{like} & \g $\mathcal{A}_{\mathrm{comp}}$ & \g $0.20$ & \g $0.17$ & \g $0.12$ & \g $0.10$ & \g $0.10$ \\
 &  & $\mathcal{A}_{\mathrm{exist}}$ & $0.06$ & $0.20$ & $0.12$ & $0.11$ & $0.10$ \\
\cmidrule(lr){2-8}
 & \multirow{2}{*}{swap} & \g $\mathcal{A}_{\mathrm{comp}}$ & \g $0.16^{*}$ & \g $0.12$ & \g $0.09$ & \g $0.04$ & \g $0.07$ \\
 &  & $\mathcal{A}_{\mathrm{exist}}$ & $0.10$ & $0.14$ & $0.10$ & $0.14$ & $0.14$ \\
\cmidrule(lr){2-8}
 & \multirow{2}{*}{obf} & \g $\mathcal{A}_{\mathrm{comp}}$ & \g $0.17$ & \g $0.09$ & \g $0.15$ & \g $0.07$ & \g $0.11$ \\
 &  & $\mathcal{A}_{\mathrm{exist}}$ & $0.13$ & $0.16$ & $0.14$ & $0.16$ & $0.18$ \\
\midrule
\multirow{8}{*}{\texttt{synthetic}} & \multirow{2}{*}{real} & \g $\mathcal{A}_{\mathrm{comp}}$ & \g $0.15$ & \g $0.12$ & \g $0.16$ & \g $0.07$ & \g $0.12$ \\
 &  & $\mathcal{A}_{\mathrm{exist}}$ & $0.12$ & $0.15$ & $0.17$ & $0.19$ & $0.17$ \\
\cmidrule(lr){2-8}
 & \multirow{2}{*}{like} & \g $\mathcal{A}_{\mathrm{comp}}$ & \g $0.26$ & \g $0.23$ & \g $0.23$ & \g $0.29$ & \g $0.24$ \\
 &  & $\mathcal{A}_{\mathrm{exist}}$ & $0.21$ & $0.23$ & $0.20$ & $0.23$ & $0.18$ \\
\cmidrule(lr){2-8}
 & \multirow{2}{*}{swap} & \g $\mathcal{A}_{\mathrm{comp}}$ & \g $0.17$ & \g $0.10$ & \g $0.16$ & \g $0.13$ & \g $0.13$ \\
 &  & $\mathcal{A}_{\mathrm{exist}}$ & $0.14$ & $0.17$ & $0.17$ & $0.15$ & $0.14$ \\
\cmidrule(lr){2-8}
 & \multirow{2}{*}{obf} & \g $\mathcal{A}_{\mathrm{comp}}$ & \g $0.15$ & \g $0.12$ & \g $0.17$ & \g $0.11$ & \g $0.14$ \\
 &  & $\mathcal{A}_{\mathrm{exist}}$ & $0.18$ & $0.20$ & $0.22$ & $0.19$ & $0.20$ \\
\midrule
\multirow{8}{*}{\texttt{titanic}} & \multirow{2}{*}{real} & \g $\mathcal{A}_{\mathrm{comp}}$ & \g $0.28$ & \g $\mathbm{0.43}$ & \g $0.28$ & \g $\mathbm{0.50}$ & \g $\mathbm{0.60}$ \\
 &  & $\mathcal{A}_{\mathrm{exist}}$ & $\mathbm{0.26}$ & $0.20$ & $0.22$ & $0.24$ & $\mathbm{0.31}$ \\
\cmidrule(lr){2-8}
 & \multirow{2}{*}{like} & \g $\mathcal{A}_{\mathrm{comp}}$ & \g $0.26$ & \g $\mathbm{0.35}$ & \g $0.28$ & \g $0.33$ & \g $\mathbm{0.40}$ \\
 &  & $\mathcal{A}_{\mathrm{exist}}$ & $0.23$ & $0.26$ & $0.13$ & $0.23$ & $0.18$ \\
\cmidrule(lr){2-8}
 & \multirow{2}{*}{swap} & \g $\mathcal{A}_{\mathrm{comp}}$ & \g $0.11$ & \g $0.21^{*}$ & \g $0.22^{*}$ & \g $0.13^{*}$ & \g $0.12^{*}$ \\
 &  & $\mathcal{A}_{\mathrm{exist}}$ & $0.05$ & $0.06$ & $0.10$ & $0.08$ & $0.04$ \\
\cmidrule(lr){2-8}
 & \multirow{2}{*}{obf} & \g $\mathcal{A}_{\mathrm{comp}}$ & \g $0.14$ & \g $0.20$ & \g $0.20$ & \g $0.17$ & \g $0.07$ \\
 &  & $\mathcal{A}_{\mathrm{exist}}$ & $0.11$ & $0.14$ & $0.08$ & $0.07$ & $0.14$ \\
\end{xltabular}

\end{document}